%% file: paper.tex
\documentclass[10pt]{article}

\usepackage{Sty/mcr}
\usepackage{Sty/pkg}
\usepackage{Sty/thmE}
\usepackage{Sty/algE}

\title{Safe Pattern Pruning: \\ An Efficient Approach for Predictive Pattern Mining}

\date{\today}

\author{
Kazuya Nakagawa \\
Nagoya Institute of Technology \\
\texttt{nakagawa.k.mllab.nit@gmail.com} \\
\and
Shinya Suzumura \\
Nagoya Institute of Technology \\
\texttt{suzumura.mllab.nit@gmail.com} \\
\and 
Masayuki Karasuyama \\
Nagoya Institute of Technology \\
\texttt{karasuyama@nitech.ac.jp} \\
\and 
Koji Tsuda \\
University of Tokyo \\
\texttt{tsuda@k.u-tokyo.ac.jp } \\
\and
Ichiro Takeuchi\thanks{Corresponding author} \\
Nagoya Institute of Technology \\
\texttt{takeuchi.ichiro@nitech.ac.jp} \\
}

\begin{document}

\maketitle

\newpage

\begin{abstract}
 \input{abst}

 \vspace{.1in}
 \begin{center}
 {\bf Keywords}
 \end{center}
 \input{keyword}
 \vspace{.1in}
\end{abstract}

\newpage

\input{Sec/sec1}
\input{Sec/sec2}

\input{Sec/sec3}
\input{Sec/sec4}

\input{Sec/sec5}

\appendix
\input{App/appA}

\end{document}

%% file: abst.tex
In this paper we study predictive pattern mining problems where the goal is to construct a predictive model based on a subset of predictive patterns in the database. 
Our main contribution is to introduce a novel method called \emph{safe pattern pruning (SPP)} for a class of predictive pattern mining problems. 
The SPP method allows us to efficiently find a superset of all the predictive patterns in the database that are needed for the optimal predictive model. 
The advantage of the SPP method over existing boosting-type method is that the former can find the superset by a single search over the database, while the latter requires multiple searches. 
The SPP method is inspired by recent development of \emph{safe feature screening}. 
In order to extend the idea of safe feature screening into predictive pattern mining,  we derive a novel pruning rule called \emph{safe pattern pruning (SPP) rule} that can be used for searching over the tree defined among patterns in the database. 
The SPP rule has a property that,
if a node corresponding to a pattern in the database is pruned out by the SPP rule,
then it is guaranteed that all the patterns corresponding to its descendant nodes are never needed for the optimal predictive model. 
We apply the SPP method to graph mining and item-set mining problems, and demonstrate its computational advantage.

%% file: keyword.tex
Predictive pattern mining, Graph mining, Item-set mining, Sparse learning, Safe screening, Convex optimization

%% file: Sec/sec1.tex
\section{Introduction} \label{sec:introduction}
In this paper, we study predictive pattern mining. 
The goal of predictive pattern mining is discovering a set of patterns from databases that are needed for constructing a good predictive model.
Predictive pattern mining problems can be interpreted as feature selection problems in supervised machine learning tasks such as classifications and regressions.
The main difference between predictive pattern mining and ordinal feature selection is that, in the former, the number of possible patterns in databases are extremely large, meaning that we cannot naively search over all the patterns in databases.
We thus need to develop algorithms that can exploit some structures among patterns such as trees or graphs for efficiently discovering good predictive patterns. 

To be concrete, suppose that there are $D$ patterns in a database, which is assumed to be extremely large. 
For the $i$-th transaction in the database, let $z_{i1}, \ldots, z_{iD} \in \{0, 1\}$ represent the occurrence of each pattern.
We consider linear predictive model in the form of 
\begin{align}
 \label{eq:predictive-model}
 \sum_{j \in \cA} w_j z_{ij} + b, 
\end{align}
where $\cA \subseteq \{1, \ldots, D\}$ is a set of patterns that would be selected by a mining algorithm, and $\{w_j\}_{j \in \cA}$ and $b$ are the parameters of the linear predictive model.
Here, the goal is to select a set of predictive patterns in  $\cA$ and find the model parameters $\{w_j\}_{j \in \cA}$ and $b$ so that the predictive model in the form of \eq{eq:predictive-model} has good predictive ability. 

Existing predictive pattern mining studies can be categorized into two approaches.
The first approach is \emph{two-stage} approach, where a mining algorithm is used for selecting the set of patterns $\cA$ in the first stage, and the predictive model is fitted by only using the selected patterns in $\cA$ in the second stage. 
Two-stage approach is computationally efficient because the mining algorithm is run only once in the first stage.
However, two-stage approach is suboptimal as predictive model building procedure because it does not directly optimize the predictive model. 
The second approach is \emph{direct} approach, where a mining algorithm is integrated in a feature selection method. 
An advantage of direct approach is that a set of patterns that are useful for predictive modeling is directly searched for. 
However, the computational cost of existing direct approach is usually much greater than two-stage approach because the mining algorithm is run multiple times.
For example, in a stepwise feature selection method, the mining algorithm is run at each step in order for finding the pattern that best improves the current predictive model. 

In this paper, we study a direct approach for predictive pattern mining based on \emph{sparse modeling}. 
In the literature of machine-learning and statistics, sparse modeling has been intensively studied in the past two decades.
An advantage of sparse modeling is that the problem is formulated as a convex optimization problem, and it allows us to investigate several properties of solutions from a wide variety of perspectives \cite{hastie2015statistical}. 
In addition, many efficient solvers that can be applicable to high-dimensional problems (although not as high as the number of patterns in databases as we consider in this paper) have been developed \cite{buhlmann2011statistics}.

Predictive pattern mining algorithms based on sparse modeling have been also studied in the literature \cite{saigo2006linear,saigo2007mining,saigo2009gboost}.
All these studies rely on a technique developed in the context of boosting \cite{demiriz2002linear}. 
Roughly speaking, in each step of the boosting-type method, a feature is selected based on a certain criteria, and an optimization problem defined over the set of features selected so far is solved. 
Therefore, when the boosting-type method is used for predictive pattern mining tasks, one has to search over the database as many times as the number of steps in the boosting-type method. 

Our main contribution in this paper is to propose a novel method for sparse modeling-based predictive pattern mining. 
Denoting the set of patterns that would be used in the optimal predictive model as $\cA^*$, the proposed method can find a set of patterns $\hat{\cA} \supseteq \cA^*$, i.e., $\hat{\cA}$ contains all the predictive patterns that are needed for the optimal predictive model. 
It means that, if we solve the sparse modeling problem defined over the set of patterns $\hat{\cA}$, then it is guaranteed that the resulting predictive model is optimal. 
The main advantage of the proposed method over the above boosting-type method is that a mining algorithm is run only \emph{once} for finding the set of patterns $\hat{\cA}$. 

The proposed method is inspired by recent \emph{safe feature screening} studies \cite{ghaoui2012safe,xiang2011learning,wang2013lasso,bonnefoy2014dynamic,liu2014safe,wang2014safe,xiang2014screening,fercoq2015mind,ndiaye2015gap}. 
In ordinary feature selection problems, safe feature screening allows us to identify a set of features that would never be used in the optimal model before actually solving the optimization problem
It means that these features can be \emph{safely} removed from the training set. 
Unfortunately, however, it cannot be applied to predictive pattern mining problems because it is computationally intractable to apply safe feature screening to each of extremely large number of patterns in a database for checking whether the pattern can be safely removed out or not. 

In this paper, we develop a novel method called \emph{safe pattern pruning (SPP)}. 
Considering a tree structure defined among patterns in the database, the SPP method allows us to prune the tree in such a way that, if a node corresponding to a pattern in the database is pruned out, then it is guaranteed that all the patterns corresponding to its descendant nodes would never be needed for the optimal predictive model. 
The SPP method can be effectively used in predictive pattern mining problems because we can identify an extremely large set of patterns that are irrelevant to the optimal predictive model by exploiting the tree structure among patterns in the database,
A superset $\hat{\cA} \supseteq \cA^*$ can be obtained by collecting the set of patterns corresponding to the nodes that are not pruned out by the SPP method. 

\subsection{Notation and outline}
\label{subsec:notation-outline}
We use the following notations in the rest of the paper. 
For any natural number $n$, we define $[n] := \{1, \ldots, n\}$. 
For an $n$-dimensional vector $\bm v$ and a set $\cI \subseteq [n]$, $\bm v_{\cI}$ represents a sub-vector of $\bm v$ whose elements are index by $\cI$.
The indicator function is written as $I(\cdot)$, i.e., $I(z) = 1$ if $z$ is true, and $I(z)=0$ otherwise.
Boldface $\bm 0$ and $\bm 1$ indicate a vector of all zeros and ones, respectively. 

Here is the outline of the paper.
\S\ref{sec:preliminaries}
presents 
problem setup 
and
existing methods. 
\S\ref{sec:safe-pattern-pruning} 
describes
our main contribution 
where
we introduce safe pattern pruning (SPP) method. 
\S\ref{sec:experiments}
covers
numerical experiments
for demonstrating the advantage of the SPP method. 
\S\ref{sec:conclusions} 
concludes the paper. 

%% file: Sec/sec2.tex
\section{Preliminaries} \label{sec:preliminaries}
\begin{figure*}
\begin{center}
\begin{tabular}{cccccc}
\includegraphics[scale=0.215]{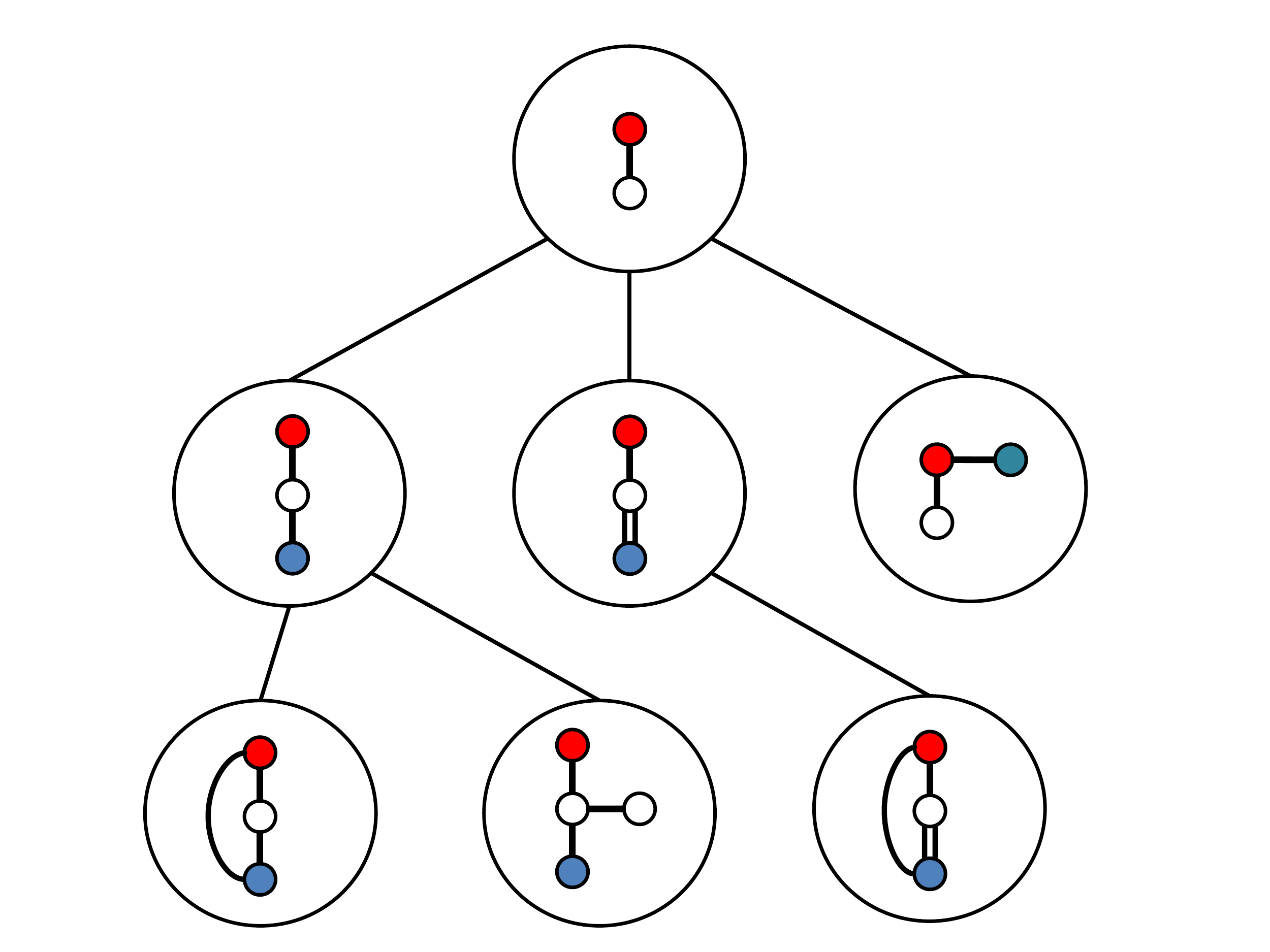} & & & & &
\includegraphics[scale=0.215]{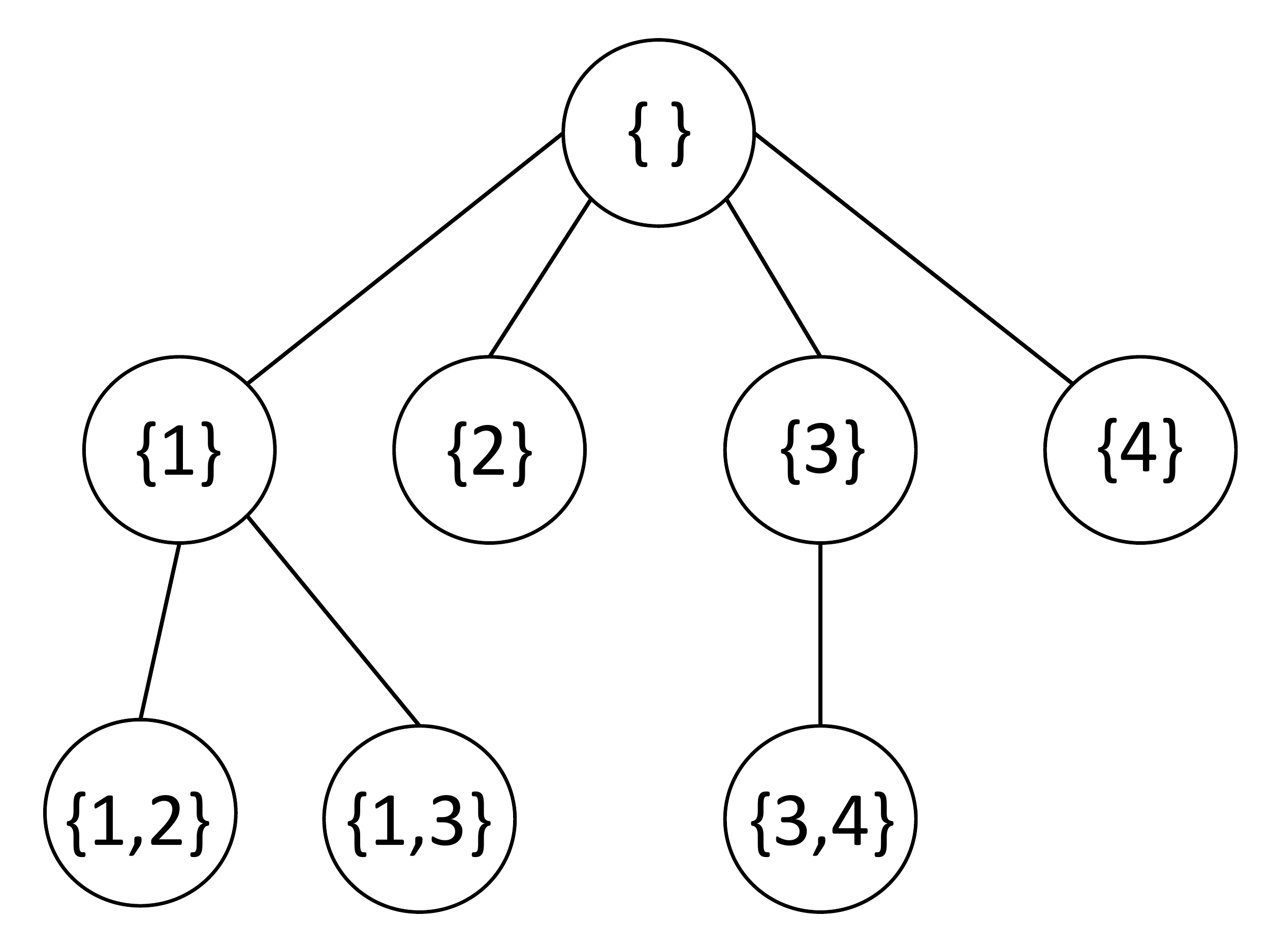} \\\\
(a) Graph mining & & & & &
(b) Item-set mining
\end{tabular}
\end{center}
\caption{Two examples of tree structures defined among patterns in databases.}
\label{fig:tree-example}
\end{figure*}

We first formulate our problem setting.

\subsection{Problem setup} \label{subsec:problem-setup}
In this paper we consider predictive pattern mining problems.
Let us consider a database with $n$ records,
and
denote the dataset as 
$\{(G_i, y_i)\}_{i \in [n]}$, 
where
$G_i$ is a labeled undirected graph in the case of graph mining,
while it is a set of items in the case of item-set mining. 
The response variable 
$y_i$
is defined
on 
$\RR$
and 
on
$\{\pm 1\}$
for
regression
and
classification problems,
respectively. 
Let
$\cT$
be the set of all patterns in the database,
and
denote its size as
$D := |\cT|$. 
For example, 
$\cT$ is the set of all possible subgraphs 
in the case of graph mining,
while
$\cT$ is the set of all possible item-sets 
in the case of item-set mining. 
Alternatively, 
$G_i$
is represented as a $D$-dimensional binary vector 
$\bm x_i \in \{0, 1\}^D$
whose $t$-th element is defined as 
\begin{align*}
 x_{it} := I(t \subseteq G_i),
 ~
 \forall t \in \cT.
\end{align*}
%
The number of patterns $D$ is extremely large in all practical pattern mining problems. 
It implies that 
any algorithms
that naively search over all $D$ patterns are computationally infeasible. 

In order to study 
both regression and classification problems
in a unified framework,
we consider the following class of convex optimization problems: 
\begin{align}
 \label{eq:general-problem}
 \min_{\bm w, b} P_\lambda(\bm w, b)
:=
\sum_{i \in [n]}
f(\bm \alpha_i^\top \bm w + \beta_i b + \gamma_i)
+ \lambda \| \bm w \|_1, 
\end{align}
where
$f: \RR \to \RR$
is a gradient Lipschitz continuous loss function
and
$\lambda > 0$
is a tuning parameter. 
We refer the problem 
\eq{eq:general-problem}
as
\emph{primal problem}
and write the optimal solution
as
$\bm w^*$. 
When
$f(z) := \frac{1}{2} z^2$
and 
$\bm \alpha_i := \bm x_i$,
$\beta_i := 1$,
$\gamma_i := -y_i$
$\forall i \in [n]$, 
the general problem
\eq{eq:general-problem}
is reduced to the following $L_1$-penalized regression problem 
defined over
$D + 1$
variables: 
\begin{align}
 \label{eq:regression-problem}
 \min_{\bm w \in \RR^D, b \in \RR}
 ~
 \frac{1}{2}
 \sum_{i \in [n]}
 (\bm x_i^\top \bm w + b - y_i)^2
 +
 \lambda
 \|\bm w\|_1. 
\end{align}
On the other hand,
when
$f(z) := \frac{1}{2} \max \{0, 1 - z\}^2$ 
and 
$\bm \alpha_i := y_i \bm x_i$,
$\beta_i := y_i$,
$\gamma_i := 0$
$\forall i \in [n]$, 
the general problem
\eq{eq:general-problem}
is reduced to the following $L_1$-penalized classification problem 
defined over
$D + 1$
variables:
\begin{align}
 \label{eq:classification-problem}
 \min_{\bm w \in \RR^D, b \in \RR}
 ~
 \frac{1}{2} \sum_{i \in [n]}
 \max \{0, 1 - y_i (\bm x_i^\top \bm w + b)\}^2
 +
 \lambda
 \|\bm w\|_1. 
\end{align}
Remembering that 
$D$ is extremely large, 
we cannot solve these $L_1$-penalized regression and classification problems 
in a standard way.

The dual problem of
\eq{eq:general-problem}
is defined as
\begin{align}
\label{eq:dual-problem}
\begin{aligned}
 \max_{\bm \theta \in \RR^n}
 ~
 D_\lambda(\bm \theta)
 :=
 - \frac{\lambda^2}{2} \|\bm \theta\|_2^2
 + \lambda \bm \delta^\top \bm \theta
 \\
 {\rm s.t.}
 ~
 \left|
 \sum_{i \in [n]} \alpha_{it} \theta_i
 \right|
 \le 1,
 \forall t \in \cT,
 \\
 \bm \beta^\top \bm \theta = 0, 
 ~
 \theta_i \ge \veps,~
 \forall i \in [n], 
\end{aligned}
\end{align}
where
$\bm \delta = \bm y$, $\veps = -\infty$
for regression problem in \eq{eq:regression-problem}, 
and
$\bm \delta = \bm 1$, $\veps = 0$
for classification problem in \eq{eq:classification-problem}.
The dual optimal solution is denoted as
$\bm \theta^*$. 

The key idea for handling an extremely large number of patterns in the database is
to exploit the tree structure defined among the patterns.
\figurename~\ref{fig:tree-example}
shows tree structures for graph mining (left) and item-set mining (right). 
As shown in
\figurename~\ref{fig:tree-example},
each node of the tree corresponds to each pattern in the database.
Those trees are constructed in such a way that, 
for any pair of a node $t$
and
one of its descendant node $t^\prime$, 
they satisfy the relation
$t \subseteq t^\prime$,
i.e., 
the pattern 
$t^\prime$
is a superset of
the pattern
$t$. 
It suggests that, 
for such a pair of $t$ and $t^\prime$, 
\begin{align*}
 x_{i t^\prime} = 1
 ~\Rightarrow~
 x_{i t} = 1
 ~~~
 \forall i, 
\end{align*}
and, conversely
\begin{align*}
 x_{i t} = 0
 ~\Rightarrow~
 x_{i t^\prime} = 0
 ~~~
 \forall i.
\end{align*}

\subsection{Existing method} \label{subsec:existing-method}
To the best of our knowledge,
except for 
the boosting-type method
described in
\S\ref{sec:introduction}
and its extensions or modifications \cite{saigo2006linear,saigo2007mining,saigo2009gboost}, 
there is no other existing method
that can be used for solving the convex optimization problem 
\eq{eq:general-problem}
for predictive pattern mining problems 
defined over an extremely large number of patterns $D$.
The boosting-type method solves the dual problem
\eq{eq:dual-problem}.
The difficulty in the dual problem is that
there are extremely large number of constraints
in the form of
$| \sum_{i \in [n]} \alpha_{it} \theta_i | \le 1, \forall t \in \cT$. 
Starting from the optimization problem 
\eq{eq:dual-problem} 
without these constraints, 
in each step of the boosting-type method, 
the most violating constraint is added to the problem, 
and 
an optimization problem 
only with
the constraints added so far
is solved. 
In optimization literature, this approach is generally known as
the \emph{cutting-plane} method, for which 
its effectiveness has been also shown in some machine learning problems 
(e.g., \cite{joachims2006training}).  
The key computational trick used by \cite{saigo2006linear,saigo2007mining,saigo2009gboost} is that, 
for finding the most violating constraint in each step, 
it is possible to efficiently search over the database 
by using a certain pruning strategy in the tree
as depicted in \figurename~\ref{fig:tree-example}. 
This method is terminated when there is no violating constraints
in the database.
 
In each single step of the boosting-type method,
one first has to search over the database by a mining algorithm, 
and then run a convex optimization solver for the problem with the newly added constraint. 
Boosting-type method is computationally expensive
because these steps must be repeated
until all the constraints
corresponding to all the predictive patterns in $\cA^*$
are added. 
In the next section,
we propose a novel method called \emph{safe pattern pruning}, 
by which
the optimal model is obtained 
by
\emph{a single search} over the database 
and
\emph{a single run} of convex optimization solver.

%% file: Sec/sec3.tex
\section{Safe pattern pruning} \label{sec:safe-pattern-pruning}
In this section, we present our main contribution. 

\subsection{Basic idea} \label{subsec:basic-idea}
It is well known that 
$L_1$ penalization
in \eq{eq:general-problem} 
makes the solution
$\bm w^*$
sparse,
i.e., 
some of its element 
would be zero. 
The set of patterns which has non-zero coefficients are called 
\emph{active} 
and
denoted as
$\cA^* \subseteq \cT$,
while the rest of the patterns are called
\emph{non-active}. 
%
A nice property of sparse learning is that
the optimal solution does not depend on any non-active patterns. 
It means that,
after some non-active patterns are removed out from the dataset,
the same optimal solution can be obtained. 
The following lemma formally states this well-known but important fact. 
\begin{lemm}
 \label{lemm:active}
 Let $\hat{\cA}$ be a set such that
 $\cA^* \subseteq \hat{\cA} \subseteq \cT$, and
 $P_\lambda^{\hat{\cA}} (\bm w_{\hat{\cA}}, b)$ be the objective function of  
 (\ref{eq:general-problem}) in which 
 $\bm w_{\cT \setminus \hat{\cA}} = \bm 0$ is substituted:  
 \begin{align} \label{eq:lem1-a}
  P_\lambda^{\hat{\cA}} (\bm w_{\hat{\cA}}, b) :=
  \sum_{i \in [n]}
  f(\bm \alpha_{\hat{\cA}, i}^\top \bm w_{\hat{\cA}} + \beta_i b + \gamma_i) + \lambda \|\bm w_{\hat{\cA}}\|_1. 
 \end{align}
 Then,
 the optimal solution of the original problem \eq{eq:general-problem} 
 is given by
 \begin{align*} 
  (\bm w^*_{\hat{\cA}}, b) &= \arg \min_{\bm w_{\hat{\cA}} \in \RR^{|\hat{\cA}|}, b \in \RR}
  ~
  P_\lambda^{\hat{\cA}} (\bm w_{\hat{\cA}}, b), \\
  \bm w^*_{\cT \setminus \hat{\cA}} &= \bm 0. 
 \end{align*}
\end{lemm}
Lemma~\ref{lemm:active}
indicates that,
if we have a set of patterns
$\hat{\cA} \supseteq \cA^*$, 
we have only to solve a smaller optimization problem defined only with the set of patterns in $\hat{\cA}$.
It means that,
if such an $\hat{\cA}$ is available, 
we do not have to work with extremely large number of patterns in the database.

In the rest of this section, 
we propose a novel method for finding such a set of patterns 
$\hat{\cA} \supseteq \cA^*$
by searching over the database only once. 
Specifically, 
we derive a novel pruning condition 
which has a property that, 
if the condition is satisfied at a certain node,
then
all the patterns corresponding to its descendant nodes and the node itself are 
guaranteed to be non-active. 
After traversing the tree,
we simply define 
$\hat{\cA}$
be the set of nodes which are not pruned out. 
Then,
it is guaranteed that 
$\hat{\cA}$
satisfies
the condition in Lemma~\ref{lemm:active}. 
The proposed method is inspired by recent studies on safe feature screening. 
We thus call our new method as \emph{safe pattern pruning (SPP)}.

\subsection{Main theorem for safe pattern pruning} \label{subsec:safe-pattern-pruning}
The following theorem
provides
a specific pruning condition
that can be used together with any search strategies on a tree. 
%
%
Let $\cT_{\rm sub}(t) \subseteq \cT$ be a set of nodes in a subtree of
$\cT$ having $t$ as a root node and containing all descendant nodes of $t$.
We derive a condition for safely
screening the entire $\cT_{\rm sub}(t)$ out, which is computable at the node $t$
without traversing the descendant nodes.
%
This means that, 
our rule, called \emph{safe pattern pruning rule},
tells us
whether a pattern $t^\prime$ $\in \cT_{\rm sub}(t)$ has a chance to be active or not 
based on the information available at the root node of the subtree $t$. 
%
%
An important consequence of the condition below is that
if the condition holds, i.e., any 
$t^\prime \in \cT_{\rm sub}(t)$ cannot be active, 
then we can stop searching over the subtree (pruning the subtree).

\begin{theo} [Safe pattern pruning (SPP) rule]\label{theo:main}
 ~\\
 Given an arbitrary primal feasible solution
 $(\tilde{\bm w}, \tilde{b})$
 and
 an arbitrary dual feasible solution 
 $\tilde{\bm \theta}$, 
 for any node $t^\prime \in \cT_{\rm sub}(t)$,
 the following
 safe pattern pruning criterion (SPPC)
 provides a rule 
 \begin{align*}
  {\rm SPPC}(t)
  :=
  u_t + r_\lambda \sqrt{v_t} < 1
  ~\Rightarrow~
  w^*_{t^\prime} = 0,
 \end{align*}
 where
 \begin{align*}
  u_t := \max \left\{
  \sum_{i: \beta_i \tilde{\theta}_i > 0} \alpha_{it} \tilde{\theta}_i,~-\sum_{i: \beta_i \tilde{\theta}_i < 0} \alpha_{it} \tilde{\theta}_i
  \right\},
  ~
  v_t := \sum_{i \in [n]} \alpha_{i t}^2, 
 \end{align*}
 for $t \in [D]$, 
 and
 \begin{align*}
  r_\lambda := 
  \frac{ \sqrt{2 (P_\lambda(\tilde{\bm w}, \tilde{b}) - D_\lambda(\tilde{\bm \theta}) ) } }{\lambda}. 
 \end{align*}
\end{theo}
\noindent
The proof of
Theorem~\ref{theo:main}
is presented in \S\ref{subsec:proof}.

${\rm SPPC}(t)$
depends on three scalar quantities 
$u_t$, $v_t$ and $r_\lambda$.
The first two quantities 
$u_t$
and 
$v_t$
are obtained 
by using information on the pattern $t$,
while 
the third quantity 
$r_\lambda$
does not depend on $t$. 
Noting that all these three quantities are non-negative, 
the SPP rule would be more powerful
(have more chance to prune the subtree)
if these three quantities are smaller. 
The following corollary is the consequence of the simple fact that the first two quantities 
$u_t$ and $v_t$
at a descendant node are smaller than those at its ancestor nodes. 
\begin{coro} \label{coro:getting-tighter}
 For any node $t^\prime \in \cT_{\rm sub}(t)$,
 \begin{align*}
  {\rm SPPC}(t) \ge {\rm SPPC}(t^\prime)
 \end{align*}
\end{coro}
The proof of Corollary~\ref{coro:getting-tighter} is presented in Appendix.
This corollary suggests that 
the SPP rule would be more powerful
at deeper nodes. 

The third quantity
$r_\lambda$
represents the goodness of the pair of primal and dual feasible solutions 
measured by the \emph{duality gap},
the difference between the primal and dual objective values. 
It means that,
if sufficiently good pair of primal and dual feasible solutions are available,
the SPP rule would be powerful. 
We will discuss
how to obtain good feasible solutions
in \S\ref{subsec:practical-considerations}.

\subsection{Proof of Theorem~\ref{theo:main}} \label{subsec:proof}
In order to prove 
Theorem~\ref{theo:main},
we first clarify the condition
for any pattern $t \in \cT$
to be non-active
by the following lemma.
\begin{lemm} \label{lemm:1st}
 For a pattern $t \in \cT$, 
 \begin{align*}
  \left| \sum_{i \in [n]} \alpha_{it} \theta^*_i \right| < 1
  ~\Rightarrow~
  w^*_t = 0. 
 \end{align*}
\end{lemm}
Proof of Lemma~\ref{lemm:1st} is presented in Appendix. 
Lemma~\ref{lemm:1st}
indicates that, 
if
an upper bound of 
$| \sum_{i \in [n]} \alpha_{i t} \theta_i^* |$
is smaller than 1,
then 
we can guarantee that
$w^*_{t} = 0$. 
%
In what follows,
we actually show that
SPPC($t$)
is an upper bound of 
$| \sum_{i \in [n]} \alpha_{i t^\prime} \theta_i^* |$
for $\forall t^\prime \in \cT_{\rm sub}(t)$.

In order to derive an upper bound of 
$| \sum_{i \in [n]} \alpha_{i t} \theta_i^* |$,
we use a technique
developed in a recent safe feature screening study \cite{ndiaye2015gap}. 
The following lemma states that, 
based on a pair of a primal feasible solution
$(\tilde{\bm w}, \tilde{b})$
and 
a dual feasible solution
$\tilde{\bm \theta}$, 
we can find a ball in the dual solution space $\RR^n$
in which 
the dual optimal solution 
$\bm \theta^*$
exists. 
\begin{lemm} [Theorem 3 in \cite{ndiaye2015gap}] \label{lemm:2nd}
 Let
 $(\tilde{\bm w}, \tilde{b})$
 be an arbitrary primal feasible solution,
 and
 $\tilde{\bm \theta}$
 be an arbitrary dual feasible solution. 
 Then,
 the dual optimal solution
 $\bm \theta^*$ 
 is within a ball in the dual solution space 
 $\RR^n$ 
 with the center
 $\tilde{\bm \theta}$
 and
 the radius
 $r_\lambda := \sqrt{2(P_\lambda(\tilde{\bm w}, \tilde{b}) - D_\lambda(\tilde{\bm \theta}))}/\lambda$. 
\end{lemm}
%
See Theorem 3 and its proof in 
\cite{ndiaye2015gap}. 
This lemma tells that,
given a pair of primal feasible and dual feasible solutions,
we can bound the dual optimal solution within a ball. 

Lemma~\ref{lemm:2nd} 
can be used for deriving an upper bound of 
$| \sum_{i \in [n]} \alpha_{i t} \theta_i^* |$.
%
Since we know that the dual optimal solution 
$\bm \theta^*$
is within the ball in Lemma~\ref{lemm:2nd},
an upper bound of any $t \in \cT$
can be obtained by solving the following convex optimization problem: 
\begin{align}
\begin{aligned}
 \label{eq:ub}
 {\rm UB}(t) := 
 \arg \max_{\bm \theta \in \RR^n} &
 ~
 \left| \sum_{i \in [n]} \alpha_{i t} \theta_i \right|
 \\
 {\rm s.t.} &
 ~
 \left\| \bm \theta - \tilde{\bm \theta} \right\|_2
 \le
 \sqrt{2(P_\lambda(\tilde{\bm w}, \tilde{b}) - D_\lambda(\tilde{\bm \theta}))}/\lambda,
 \\
 & \bm \beta^\top \bm \theta = 0.
\end{aligned}
\end{align}
Fortunately,
the convex optimization problem 
\eq{eq:ub}
can be explicitly solved
as the following lemma states.
\begin{lemm} \label{lemm:3rd}
 The solution of the convex optimization problem 
 \eq{eq:ub}
 is given as
 \begin{align*}
  {\rm UB}(t)
  =
  \left| \sum_{i \in [n]} \alpha_{i t} \tilde{\theta}_i \right|
  +
  r_\lambda \sqrt{\sum_{i \in [n]} \alpha_{i t}^2 - \frac{ (\sum_{i \in [n]} \alpha_{it} \beta_i)^2 }{ \| \bm \beta\|_2^2 }}.
 \end{align*}
\end{lemm}
Proof of Lemma~\ref{lemm:3rd} is presented in Appendix.

Although ${\rm UB}(t)$ provides a condition to screen any $t \in \cT$, 
calculating ${\rm UB}(t)$ for all $t \in \cT$ is computationally prohibiting
in our extremely high dimensional problem setting.
In the next lemma, 
we will show that 
${\rm SPPC}(t) \ge {\rm UB}(t^\prime)$
for $\forall t^\prime \in \cT_{\rm sub}(t)$,
i.e., 
${\rm SPPC}(t)$
in Theorem~\ref{theo:main}
is an upper bound of
${\rm UB}(t^\prime)$, which enables us to efficiently prune subtrees
during the tree traverse process.


\begin{lemm} \label{lemm:4th}
For any $t^\prime \in \cT_{\rm sub}(t)$,
\begin{align*}
\begin{aligned}
 {\rm UB}(t^\prime)
 \left| \sum_{i \in [n]} \alpha_{i t^\prime} \tilde{\theta}_i \right|
 +
 r_\lambda \sqrt{\sum_{i \in [n]} \alpha_{i t^\prime}^2 - \frac{ (\sum_{i \in [n]} \alpha_{it^\prime} \beta_i)^2 }{ \| \bm \beta\|_2^2 }} \\
 \le
 u_t + r_\lambda \sqrt{v_t}
 =  
 {\rm SPPC}(t).
\end{aligned}
\end{align*}
\end{lemm}

Finally,
by combining Lemmas
\ref{lemm:1st},
\ref{lemm:2nd},
\ref{lemm:3rd}
and 
\ref{lemm:4th}, 
we can prove Theorem~\ref{theo:main}. 

\noindent
{\bf Proof of Theorem~\ref{theo:main}}. 
\begin{proof}
From 
Lemmas
\ref{lemm:2nd},
\ref{lemm:3rd}
and
\ref{lemm:4th},
\begin{align}
 \label{eq:main-proof-a}
 \left| \sum_{i \in [n]} \alpha_{i t^\prime} \theta_i^* \right|
 \le
 {\rm UB}(t^\prime)
 \le
 {\rm SPPC}(t),
 ~~~
 \forall t^\prime \in \cT_{\rm sub}(t).
\end{align}
From
Lemma~\ref{lemm:1st}
and
\eq{eq:main-proof-a}, 
\begin{align*}
 {\rm SPPC}(t) 
 < 1
 ~\Rightarrow~
 w^*_{t^\prime} = 0, 
 ~~~
 \forall t^\prime \in \cT_{\rm sub}(t).
\end{align*} 
\end{proof} 

\begin{algorithm}[h]
 \caption{Regularization path computation algorithm}
 \label{alg:regularization-path}
 \begin{algorithmic}[1]
  \REQUIRE 
  $\{(G_i, y_i)\}_{i \in [n]}$,  $\{\lambda_k\}_{k \in [K]}$
  \STATE
  $\lambda_0 \lA \max_{t \in \cT} \left| \sum_{i \in [n]} x_{it} (y_i - \bar{y}) \right|$
  and
  $(\bm w_0, b_0) \lA (\bm 0, \bar{y})$
  \FOR{$k = 1, \ldots, K$}
  \STATE
  Find $\hat{\cA}(\lambda_k) \supseteq \cA^*(\lambda_k)$
  by searching over the tree with the SPP rules
  based on
  $( \bm w^*(\lambda_{k-1}), b^*(\lambda_{k-1}) )$
  and
  $ \bm \theta^*(\lambda_{k-1}) $
  as the primal and dual feasible solutions, respectively. 
  \STATE
  Solve a small optimization problems
  in \eq{eq:lem1-a}
  with $\hat{\cA} = \hat{\cA}(\lambda_{k})$, 
  and obtain 
  the primal solution
  $( \bm w^*(\lambda_{k}), b^*(\lambda_{k}) )$
  and
  the dual solution
  $ \bm \theta^*(\lambda_{k}) $.
  \ENDFOR
  \ENSURE
  $\{ (\bm w^*(\lambda_k), b^*(\lambda_k)) \}_{k \in [K]}$
  and 
  $\{ (\bm \theta^*(\lambda_k) \}_{k \in [K]}$
 \end{algorithmic}
\end{algorithm}

\subsection{Practical considerations} \label{subsec:practical-considerations}
Safe pattern pruning rule 
in Theorem~\ref{theo:main} 
depends on 
a pair of 
a primal feasible solution 
$(\tilde{\bm w}, \tilde{b})$ 
and
a dual feasible solution 
$\tilde{\bm \theta}$. 
Although the rule can be constructed from any solutions as long as they are feasible, 
the power of the rule depends on the goodness of these solutions. 
Specifically, 
the criterion 
${\rm SPPC}(t)$ 
depends on the duality gap 
$P_\lambda(\tilde{\bm w}, \tilde{b}) - D_\lambda(\tilde{\bm \theta})$ 
which would vanish when these primal and dual solutions are optimal. 
Roughly speaking,
it suggests that,
if these solutions are somewhat close to the optimal ones, 
we could expect that the SPP rule is powerful. 

\subsubsection{Computing regularization path} \label{subsubsec:regularization-path}
In practical predictive pattern mining tasks, 
we need to find a good penalty parameter 
$\lambda$ 
based on a model selection technique 
such as cross-validation. 
In model selection, 
a sequence of solutions with various different penalty parameters must be trained. 
Such a sequence of the solutions is sometimes referred to as a regularization path \cite{park2007l1}. 
Regularization path of the problem
\eq{eq:general-problem}
is usually computed from larger $\lambda$ to smaller $\lambda$
because more sparse solutions would be obtained for larger $\lambda$. 
Let us write the sequence of $\lambda$s as
$\lambda_{0} > \lambda_{1} > \ldots > \lambda_K$. 
When computing such a sequence of solutions, 
it is reasonable to use warm-start approach
where the previous optimal solution at
$\lambda_{k-1}$
is used as the initial starting point of the next optimization problem at 
$\lambda_{k}$. 
In such a situation,
we can also make use of the previous solution at 
$\lambda_{k-1}$
as the feasible solution
for the safe pattern pruning rule at
$\lambda_{k}$. 
%

In sparse modeling literature,
it is custom to start from the largest possible $\lambda$
at which the primal solution is given as
$\bm w^* = \bm 0$
and
$b^* = \bar{y}$,
where $\bar{y}$ is the sample mean of $\{y_i\}_{i \in [n]}$. 
The largest $\lambda$ is given as 
\begin{align*}
 \lambda_{\rm max}
 :=
 \max_{t \in \cT}
 \left|
 \sum_{i \in [n]}
 x_{it} (y_i - \bar{y})
 \right|. 
\end{align*}
In order to solve this maximization problem over the database,
for a node $t$ and $t^\prime \in \cT_{\rm sub}(t)$, 
we can use the following upper bound
\begin{align*}
\begin{aligned}
 &\left|
 \sum_{i \in [n]}
 x_{it^\prime} (y_i - \bar{y})
 \right|
 \\
 & ~~~~~ \le
 \max
 \left\{
 \sum_{i \mid y_i - \bar{y} > 0}
 x_{it} (y_i - \bar{y}),
 -
 \sum_{i \mid y_i - \bar{y} < 0}
 x_{it} (y_i - \bar{y})
 \right\},
\end{aligned}
\end{align*}
and this upper bound can be exploited for pruning the search over the tree.

Algorithm~\ref{alg:regularization-path} shows the entire procedure for
computing the regularization path by using the SPP rule.


%% file: Sec/sec4.tex
\section{Experiments} \label{sec:experiments}
In this section, we demonstrate the effectiveness of the proposed safe pattern pruning ({\tt SPP}) method through numerical experiments.
We compare {\tt SPP} with the boosting-based method ({\tt boosting}) discussed in \S\ref{subsec:existing-method}. 

\subsection{Experimental setup}
We considered regularization path computation scenario described in \S\ref{subsubsec:regularization-path}.
Specifically,
we computed a sequence of optimal solutions of \eq{eq:general-problem}
for a sequence of 100 penalty parameters $\lambda$
evenly allocated between $\lambda_0 = \lambda_{\rm max}$ and $0.01\lambda_0$ in logarithmic scale.
%
%
%
For solving the convex optimization problems, we used coordinate gradient descent method \cite{tseng2009coordinate}.
The optimization solver was terminated when the duality gap felled below $10^{-6}$. 
In both of {\tt SPP} and {\tt boosting}, we used warm-start approach. 
In addition, the solution at the previous $\lambda$ was also used as the feasible solution for constructing the SPP rule at the next $\lambda$. 
We used gSpan algorithm \cite{yan2002gspan} for mining subgraphs. 
We wrote all the codes (except gSpan part in graph mining experiment) in C++. 
All the computations were conducted by using a single core of an Intel Xeon CPU E5-2643 v2 (3.50GHz) with 64GB MEM.

\subsection{Graph classification/regression}
We applied {\tt SPP} and {\tt boosting} to graph classification and regression problems. 
For classification, we used {\tt CPDB} and {\tt mutagenicity} datasets,
containing $n = 648$ and $n = 4377$ chemical compounds respectively,
for which the goal is to predict whether each compound has mutagenicity or not. 
%
%
For regression, we used
{\tt Bergstrom} and {\tt Karthikeyan} datasets where the goal is to predict the
melting point of each of the $n = 185$ and $n = 4173$ chemical compounds. 
%
All datasets are downloadable from \url{http://cheminformatics.org/datasets/}.\\
We considered the cases with {\tt maxpat} $\in \{5, 6, 7, 8, 9, 10\}$, where {\tt maxpat} indicates the maximum number of edges of subgraphs we wanted to find. 

\begin{figure}[t]
\begin{center}
\subfigure[{\tt CPDB}]{
\includegraphics[clip,width=0.21\textwidth]{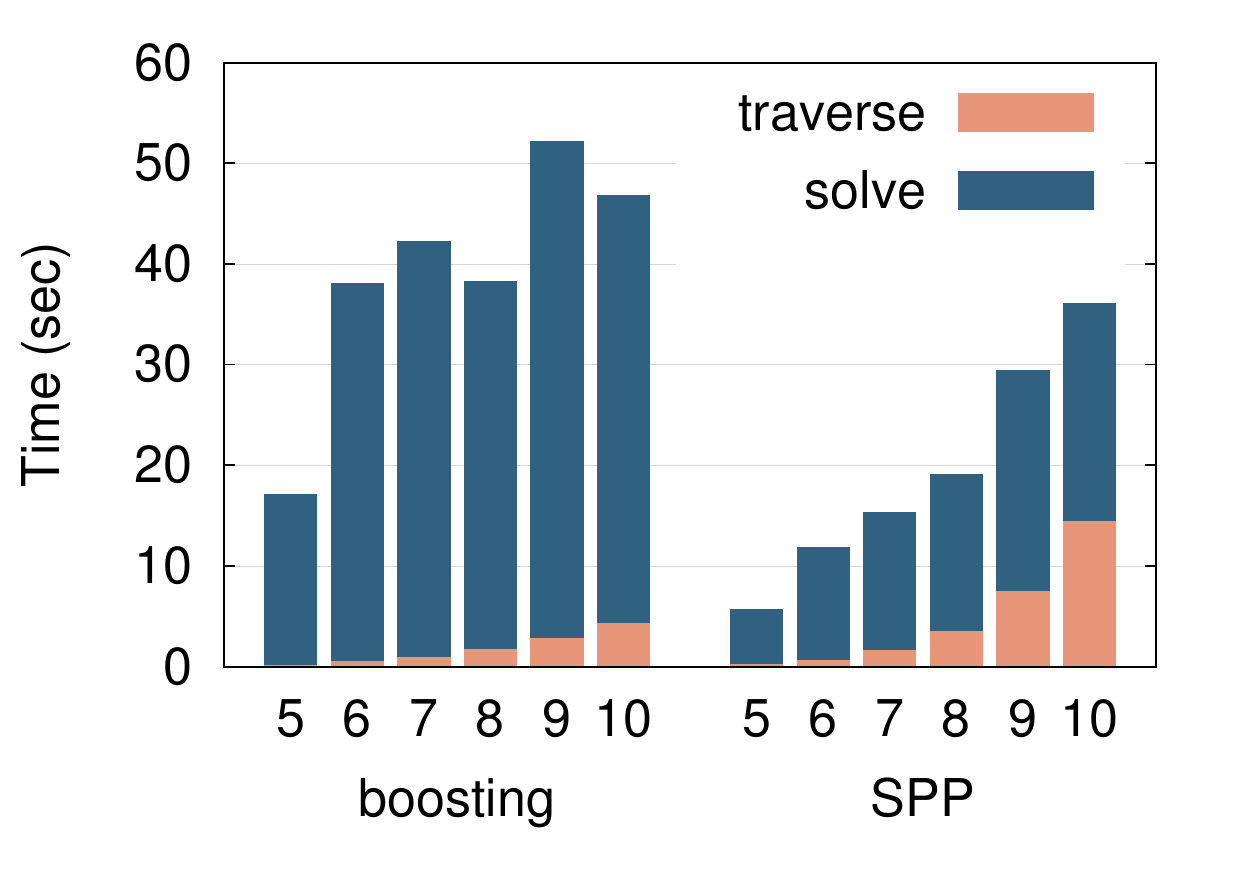}}
\subfigure[{\tt mutagenicity}]{
\includegraphics[clip,width=0.21\textwidth]{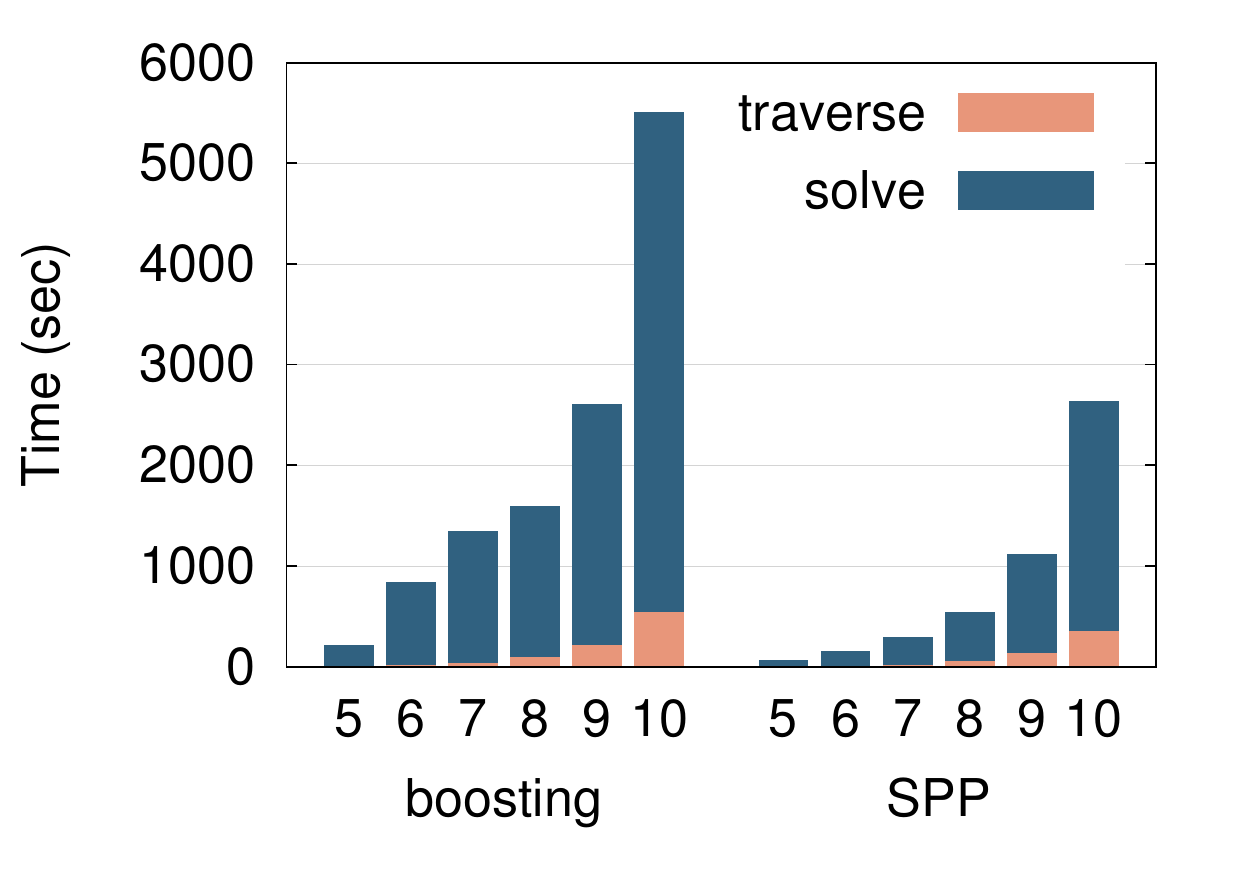}}
\subfigure[{\tt Bergstrom}]{
\includegraphics[clip,width=0.21\textwidth]{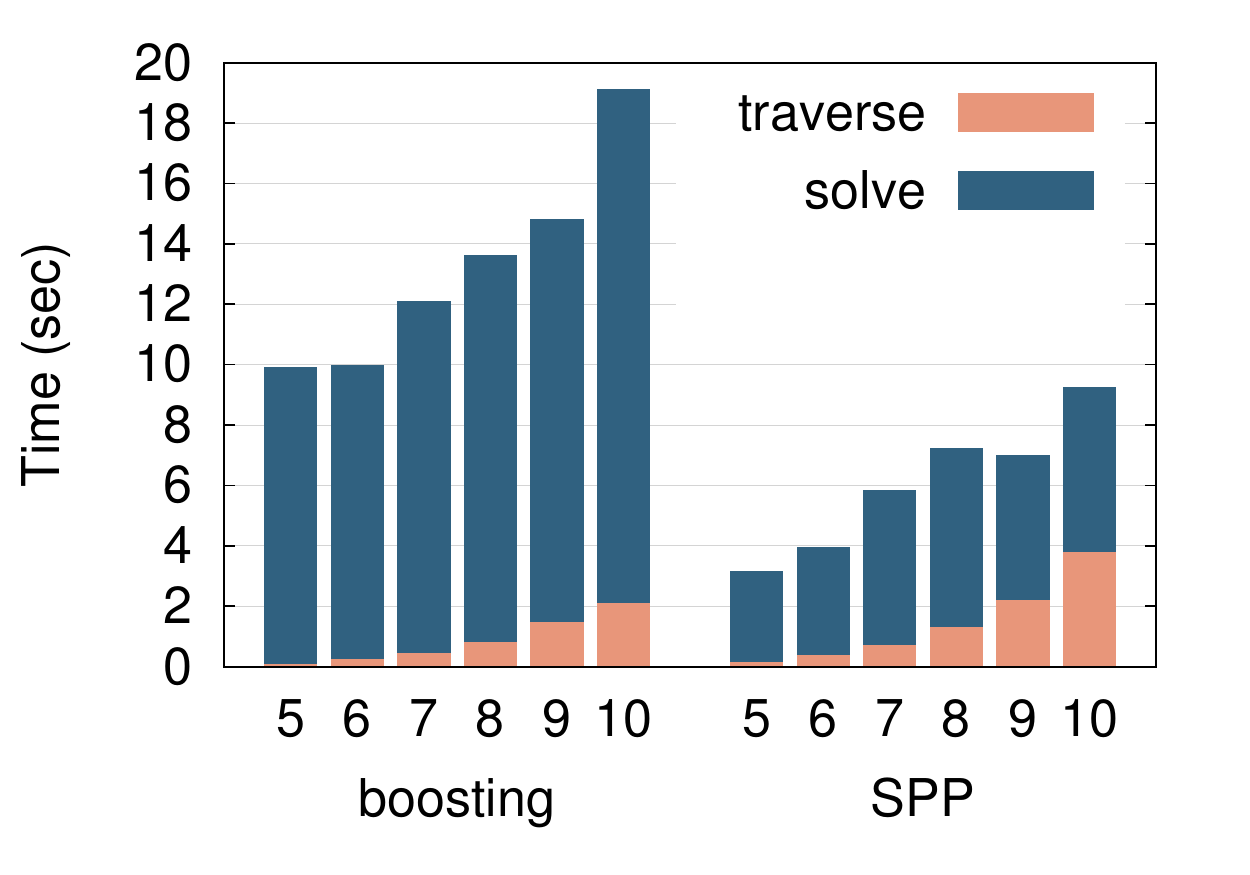}}
\subfigure[{\tt Karthikeyan}]{
\includegraphics[clip,width=0.21\textwidth]{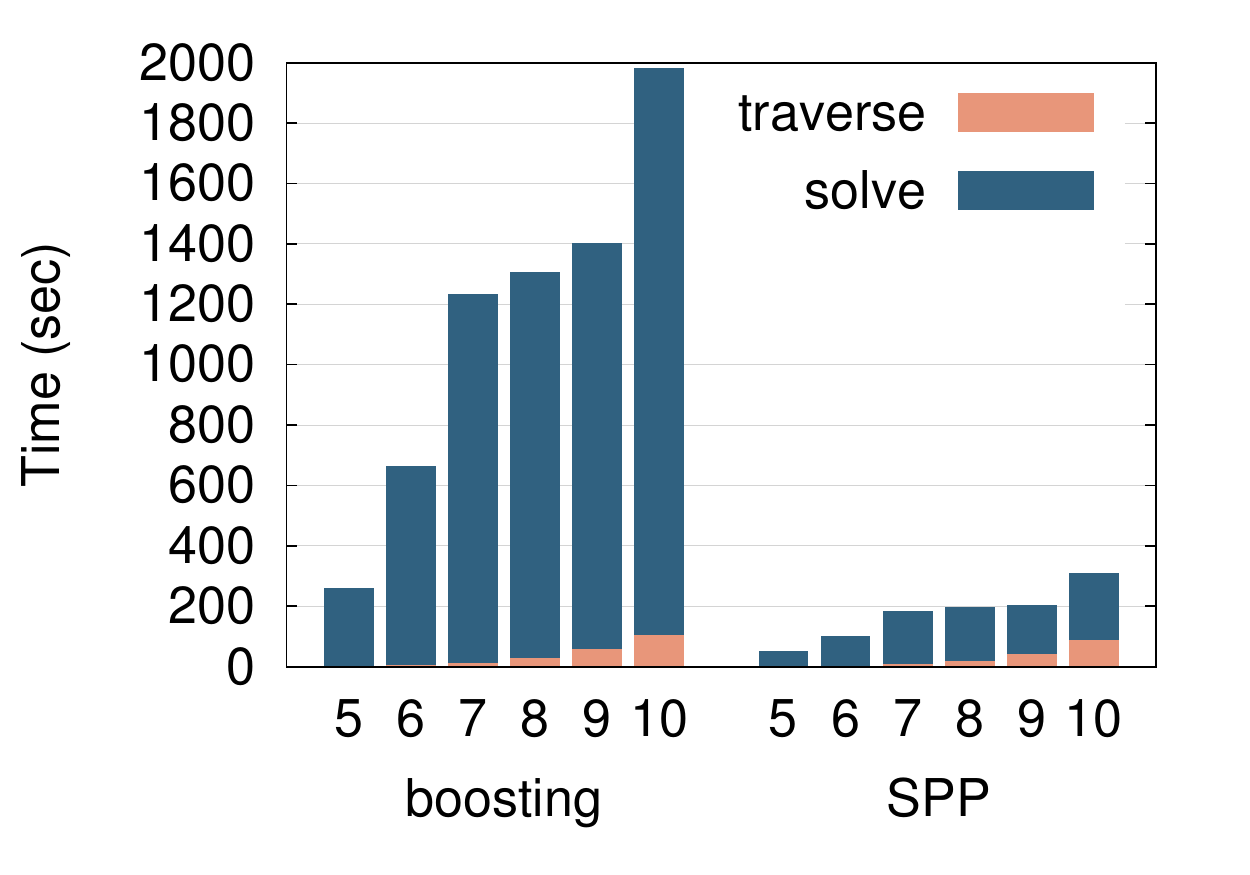}}
\end{center}
\caption{Computation time comparison for graph classification and
 regression. Each bar contains computational time taken in the tree
 traverse ({\tt traverse}) and the optimization procedure ({\tt solve}) respectively.}
\label{fig:graph_time_c}
\end{figure}

\begin{figure}
\begin{center}
\subfigure[{\tt CPDB}]{
\begin{tabular}{cccc}
\includegraphics[scale=0.33]{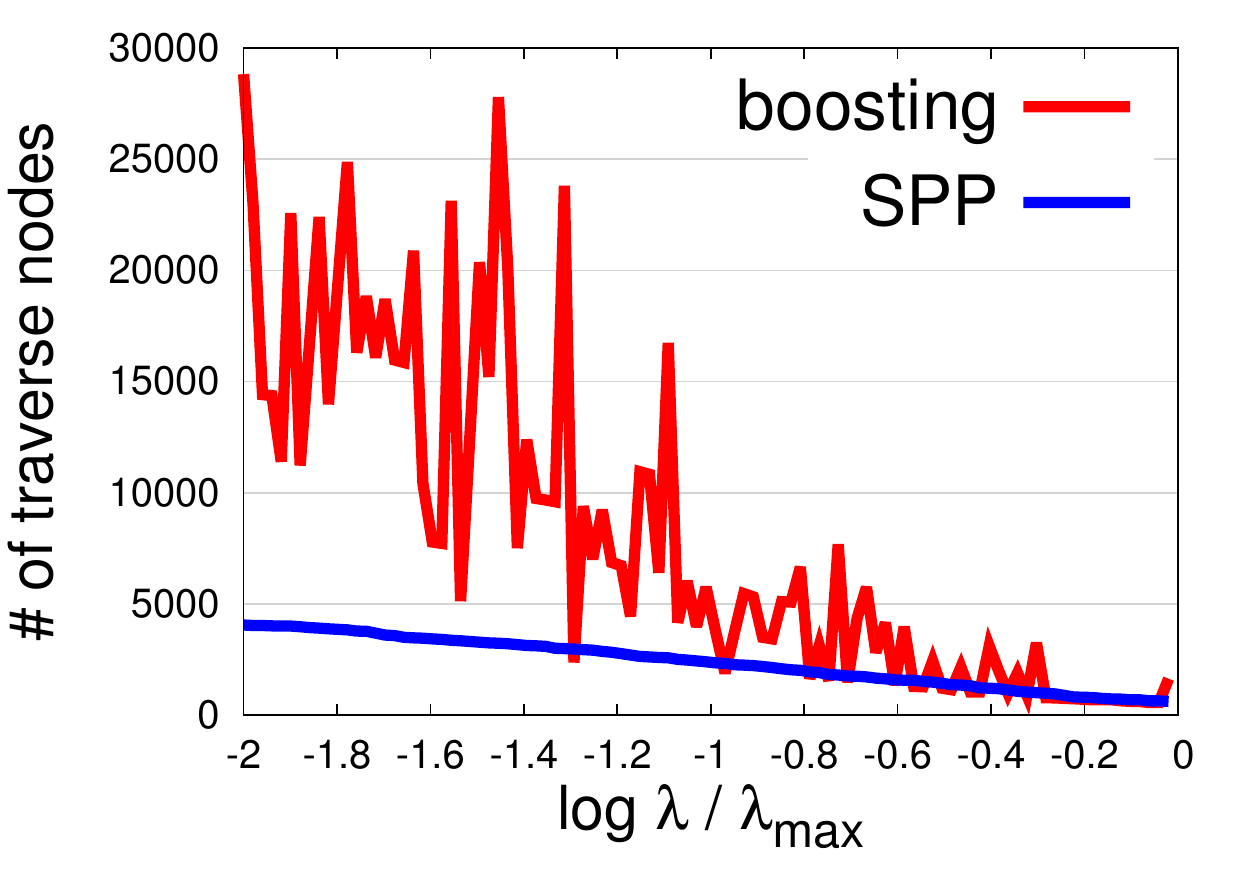} &
\includegraphics[scale=0.33]{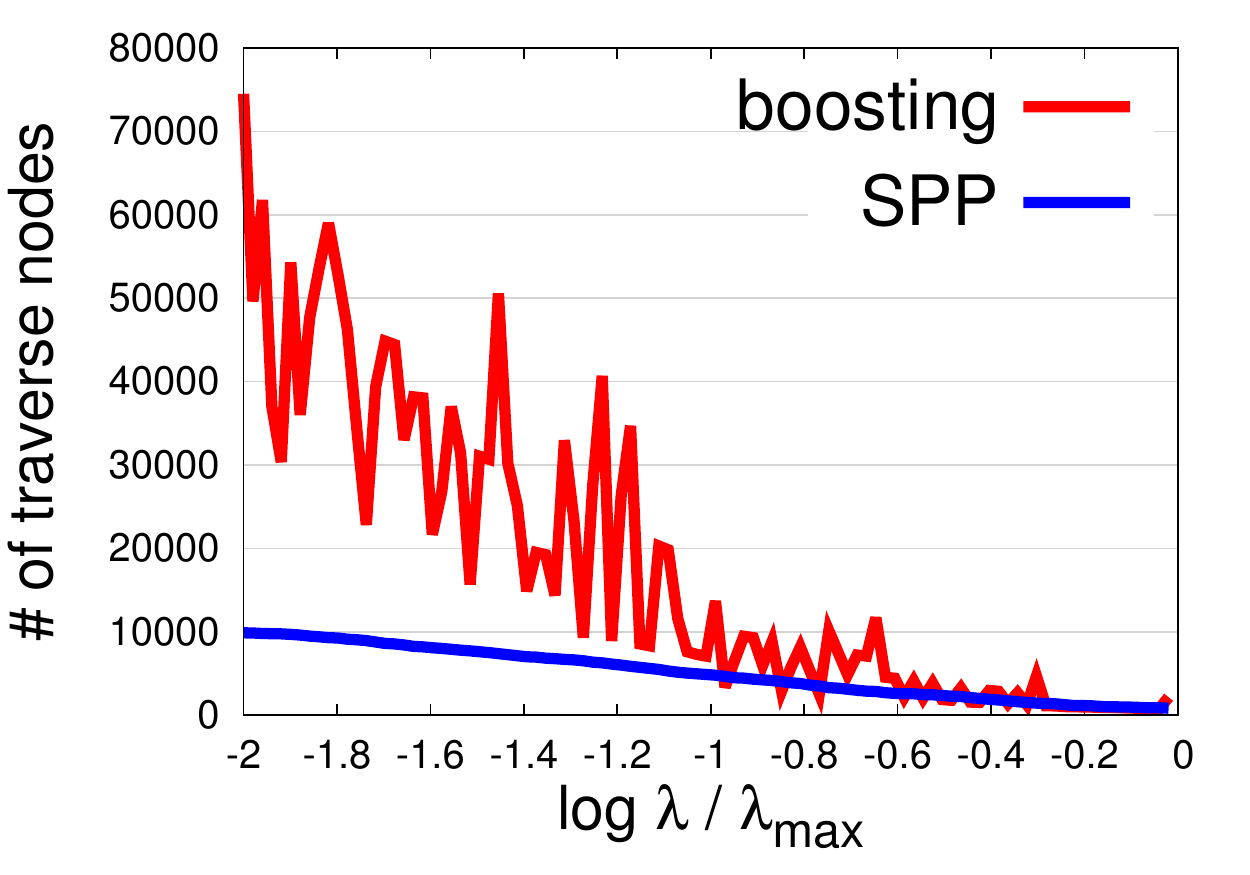} &
\includegraphics[scale=0.33]{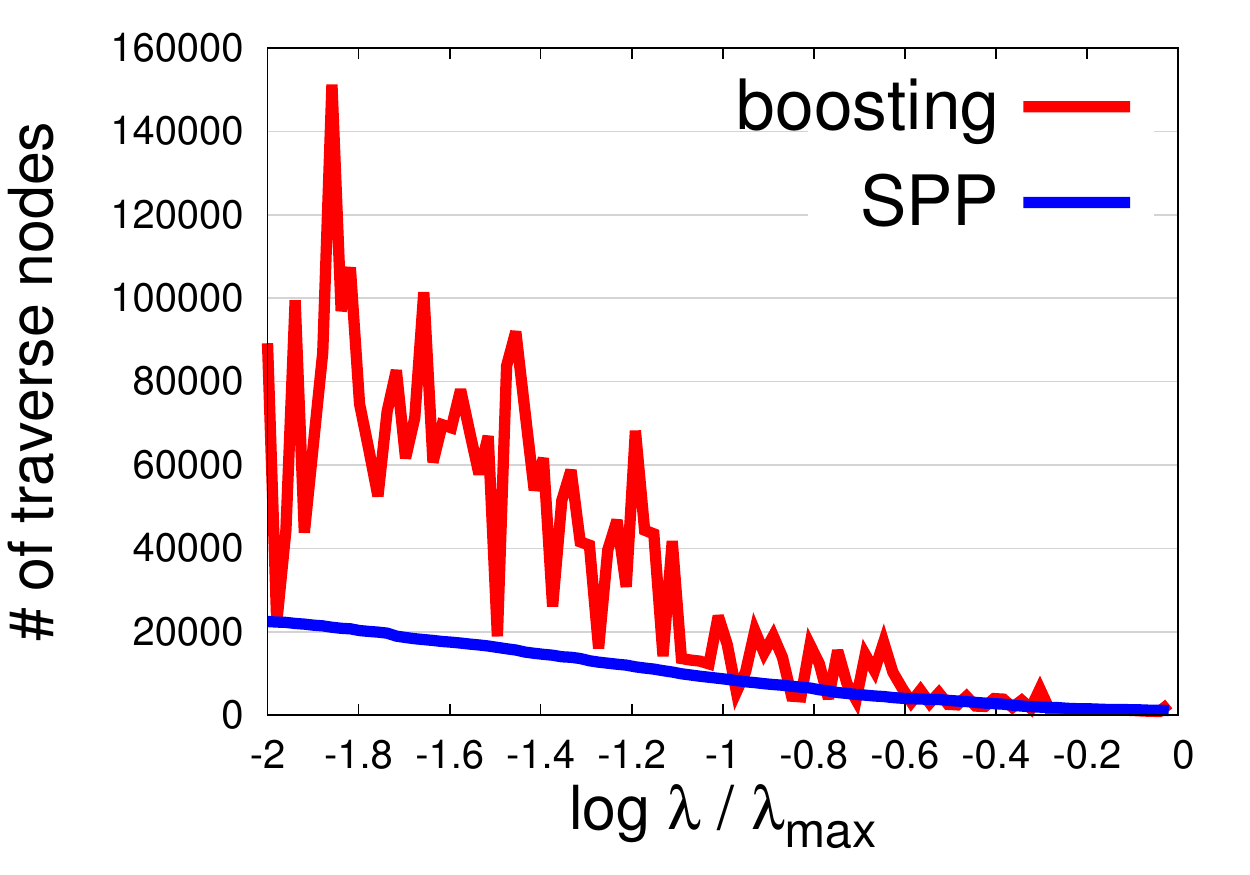} &
\includegraphics[scale=0.33]{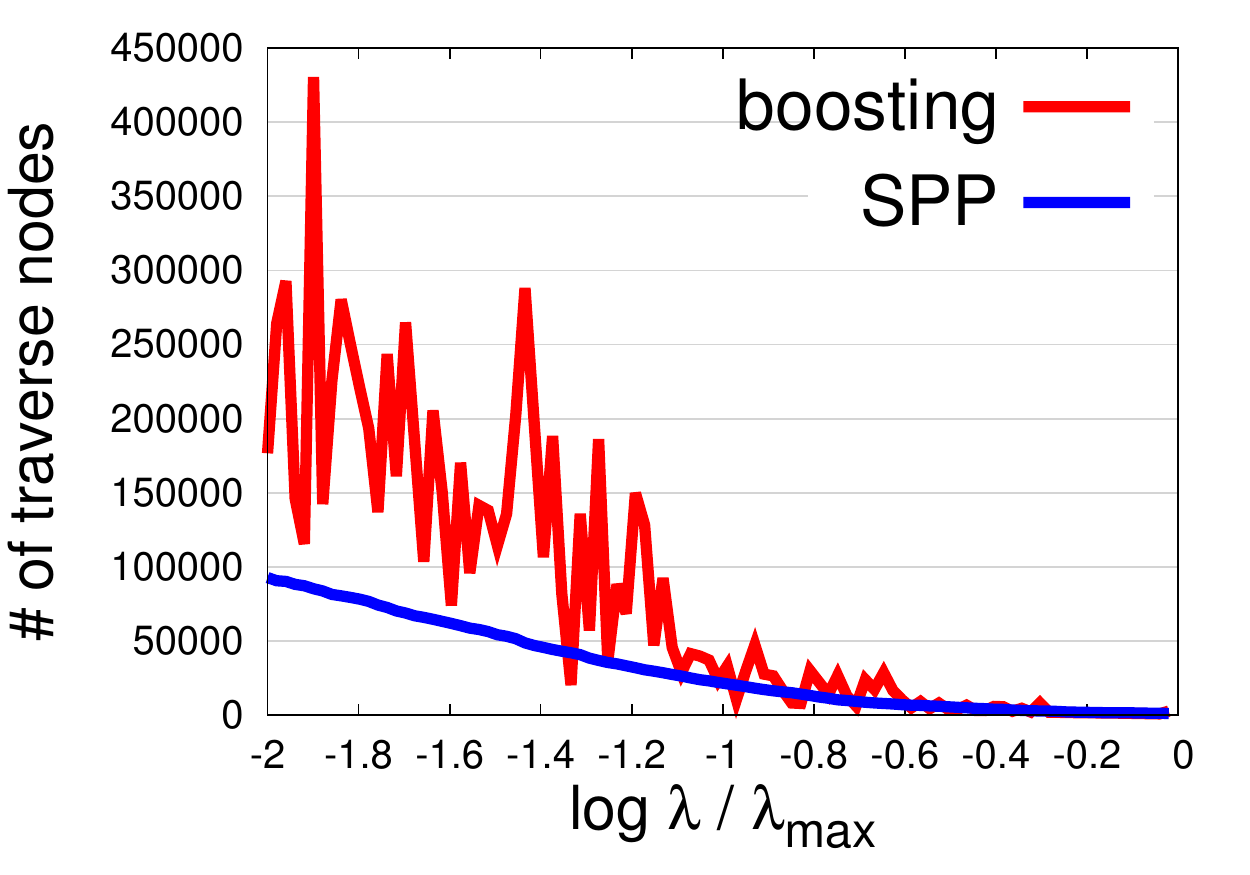} \\
(a-1) maxpat 6 &
(a-2) maxpat 7 &
(a-3) maxpat 8 &
(a-4) maxpat 10 \\\\
\end{tabular}}
\subfigure[{\tt mutagenicity}]{
\begin{tabular}{cccc}
\includegraphics[scale=0.33]{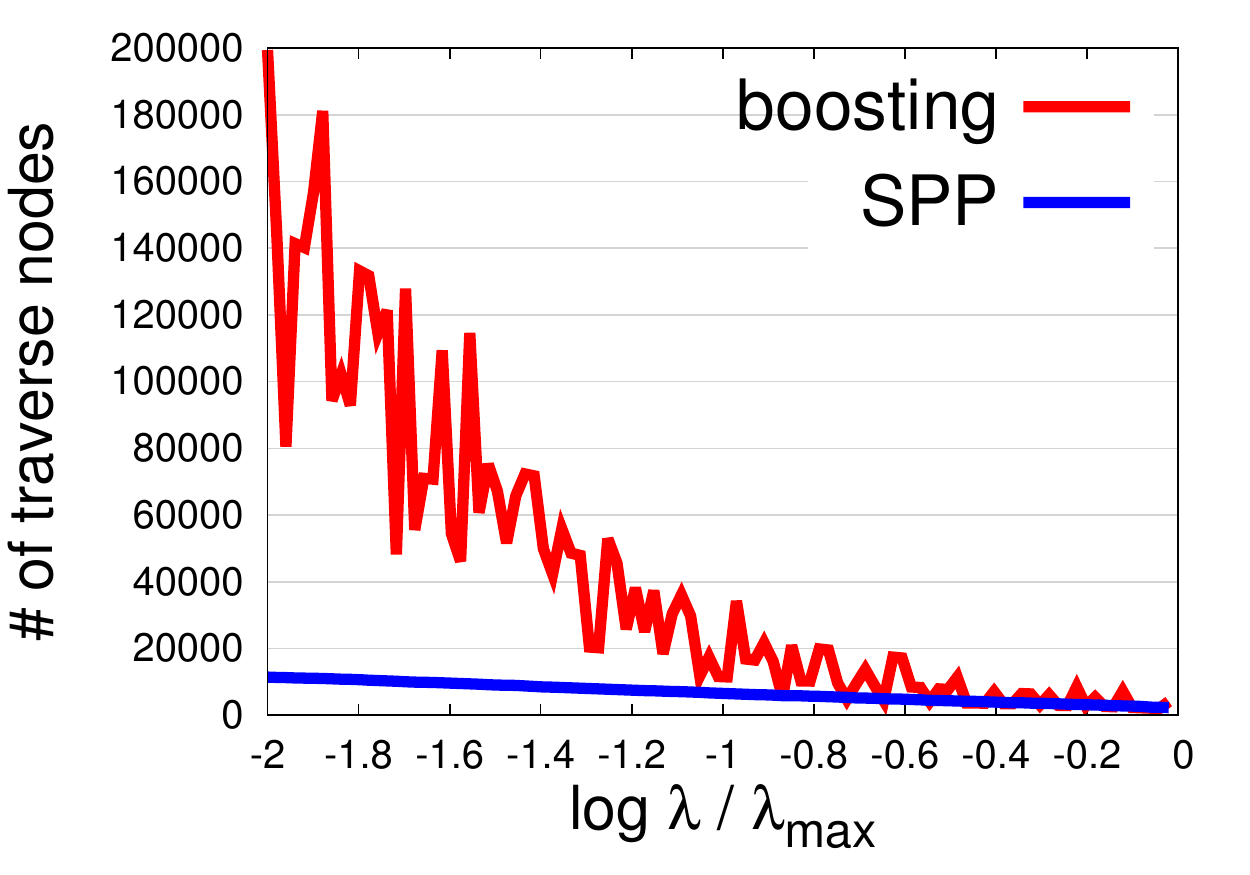} &
\includegraphics[scale=0.33]{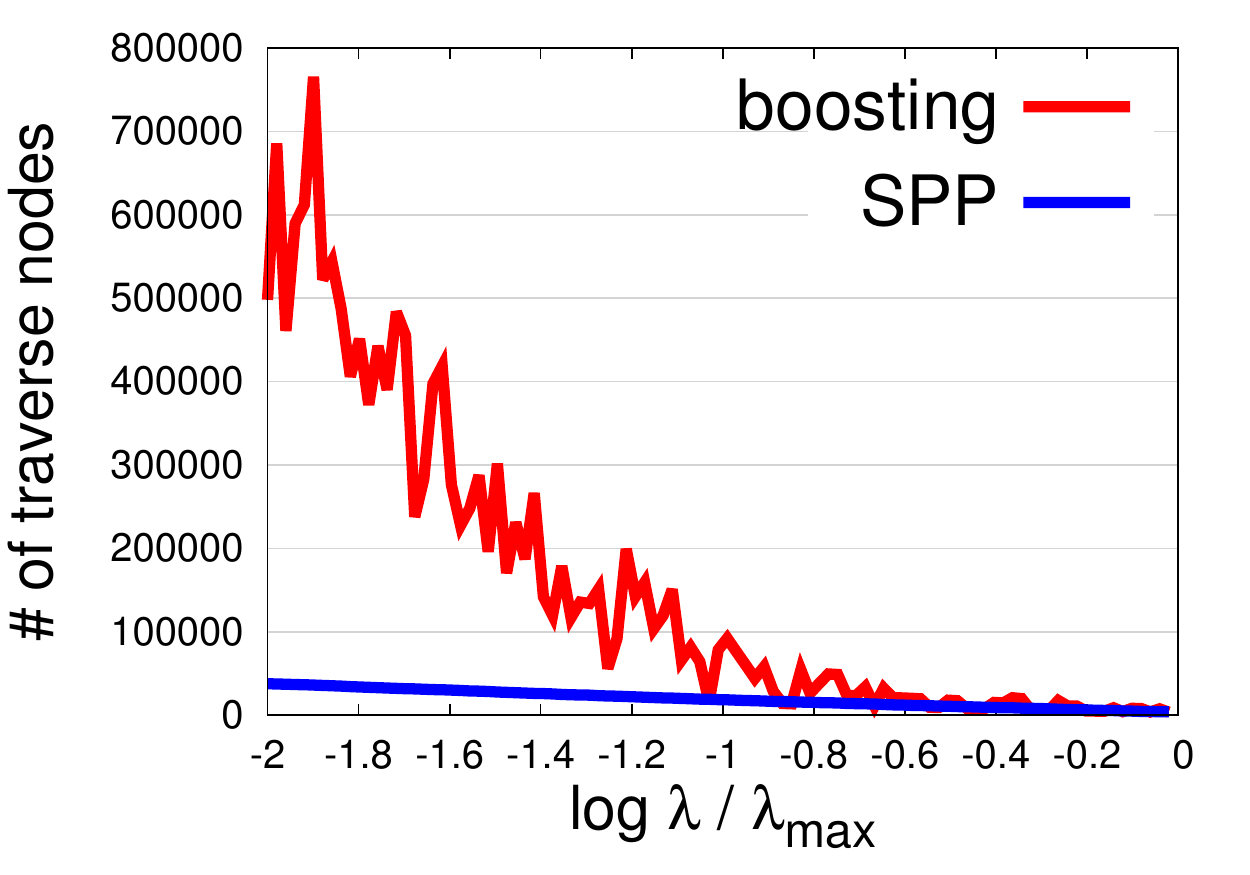} &
\includegraphics[scale=0.33]{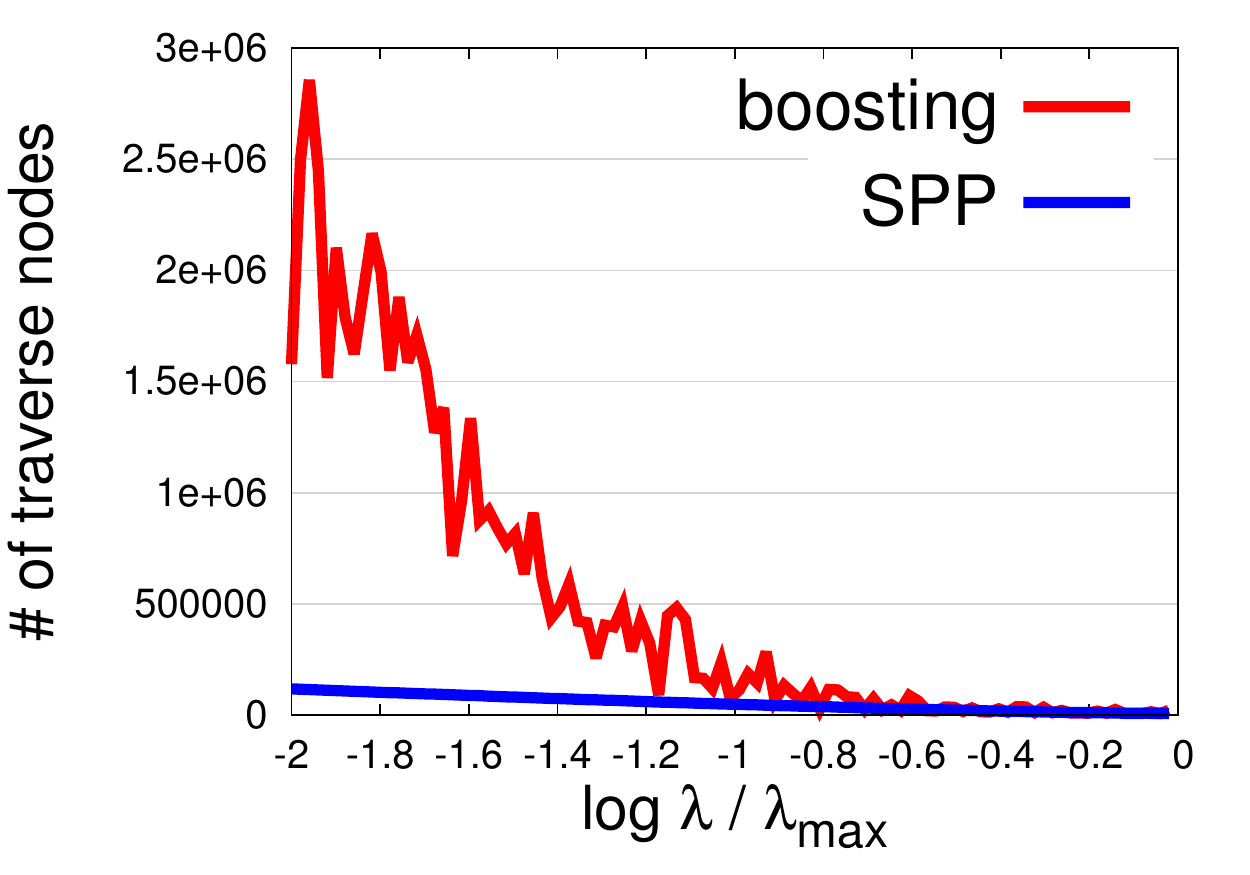} &
\includegraphics[scale=0.33]{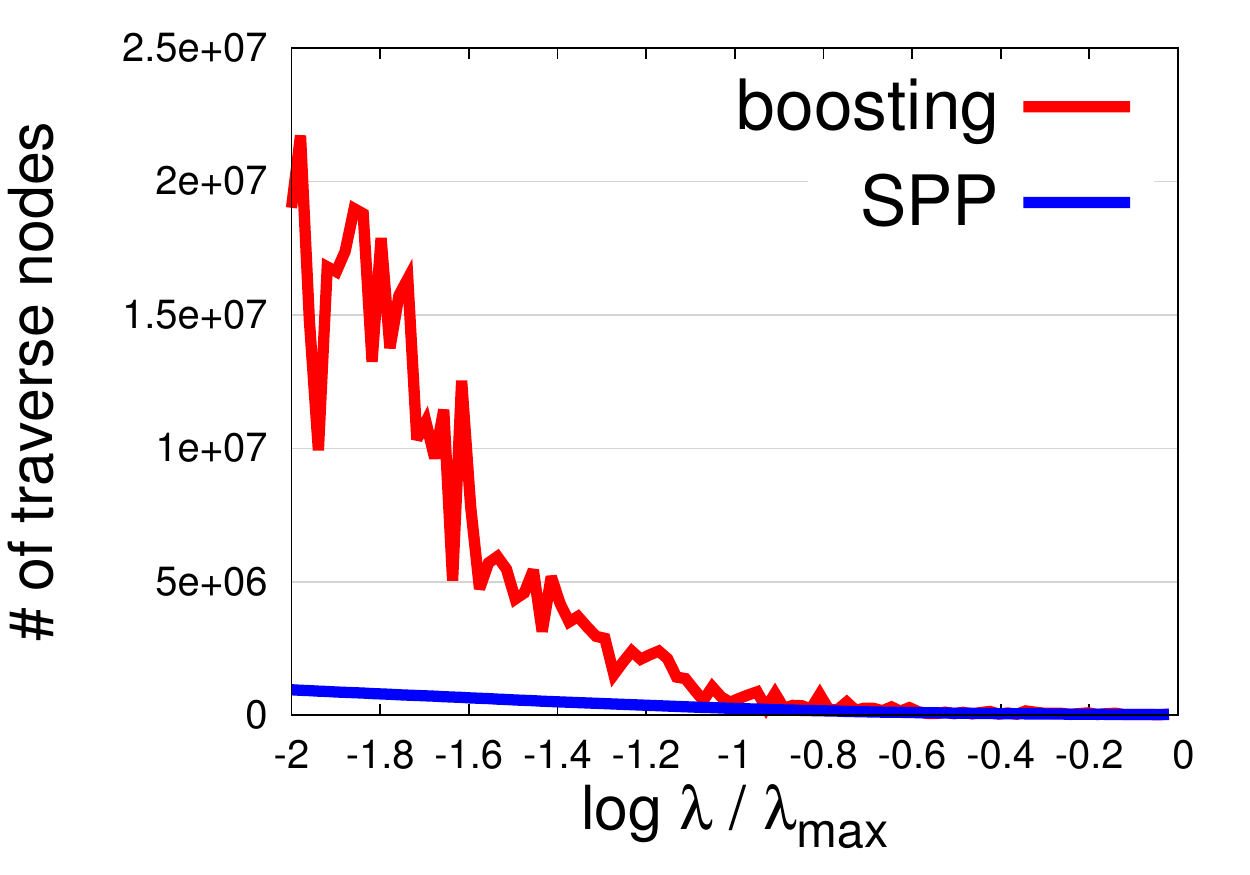} \\
(b-1) maxpat 6 &
(b-2) maxpat 7 &
(b-3) maxpat 8 &
(b-4) maxpat 10 \\\\
\end{tabular}}
\subfigure[{\tt Bergstrom}]{
\begin{tabular}{cccc}
\includegraphics[scale=0.33]{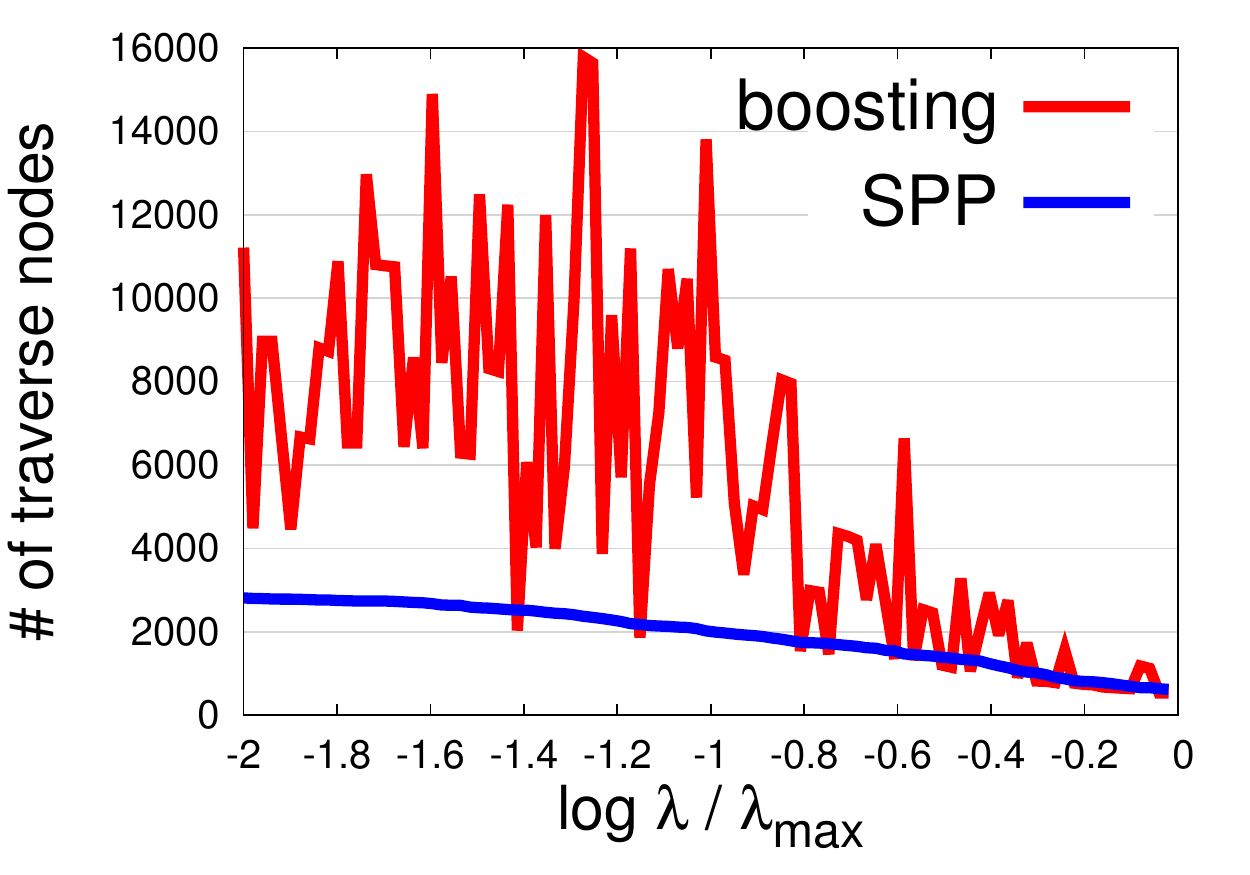} &
\includegraphics[scale=0.33]{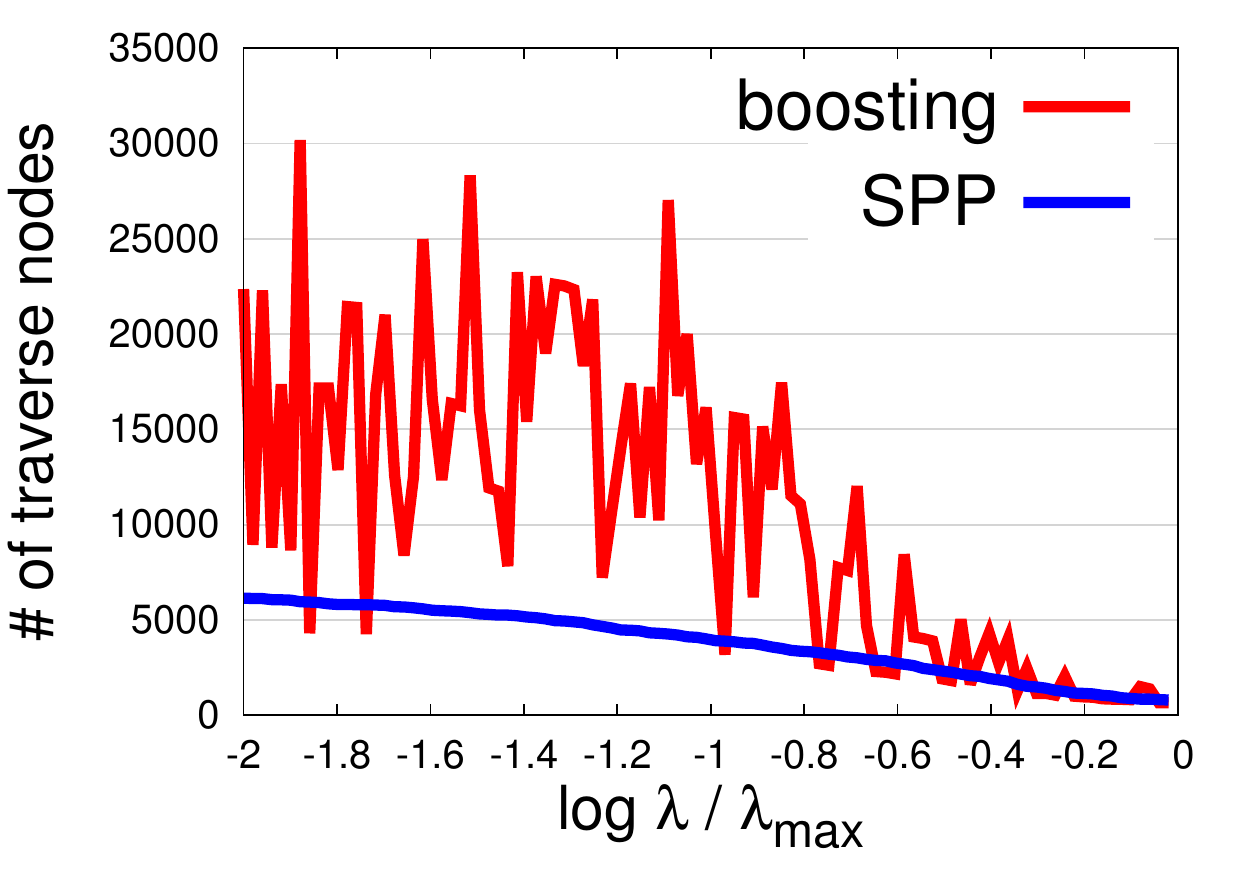} &
\includegraphics[scale=0.33]{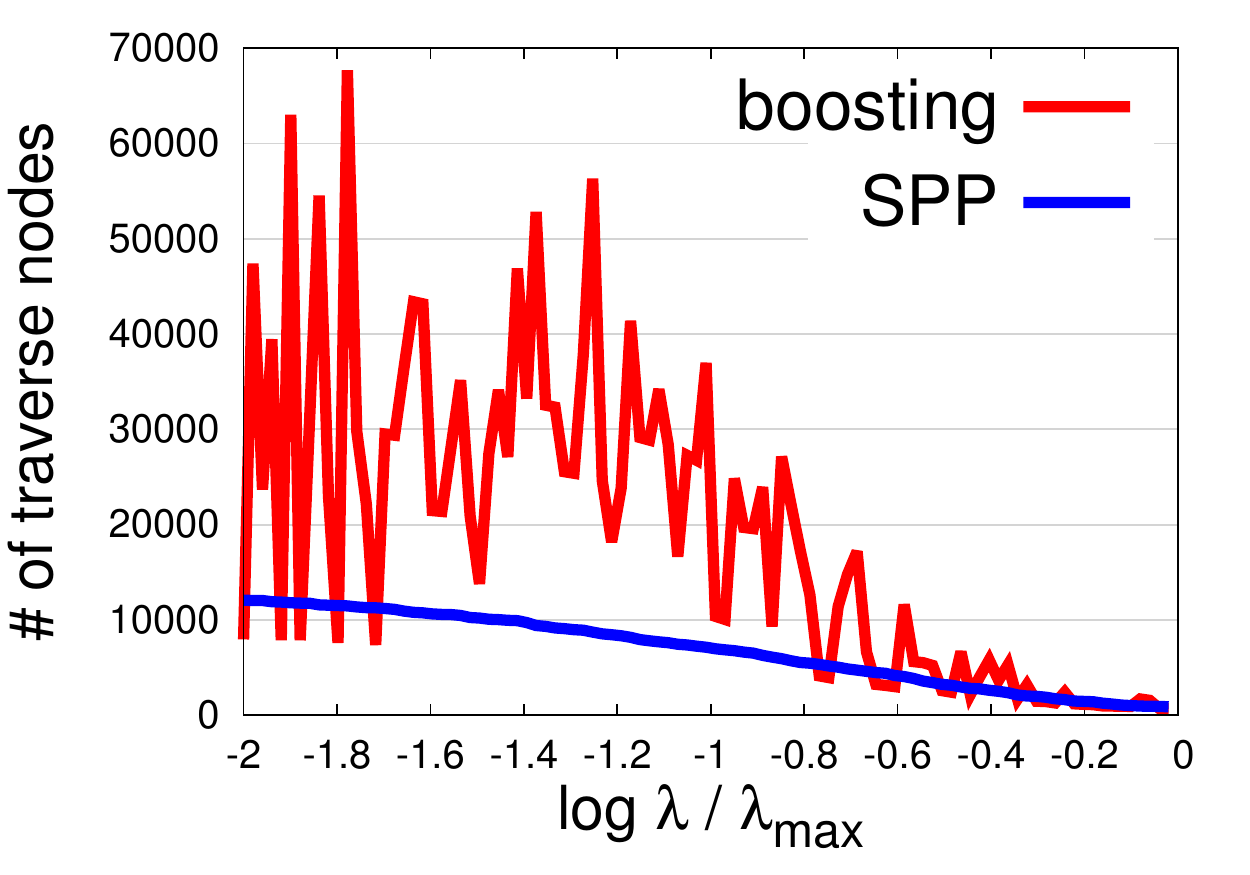} &
\includegraphics[scale=0.33]{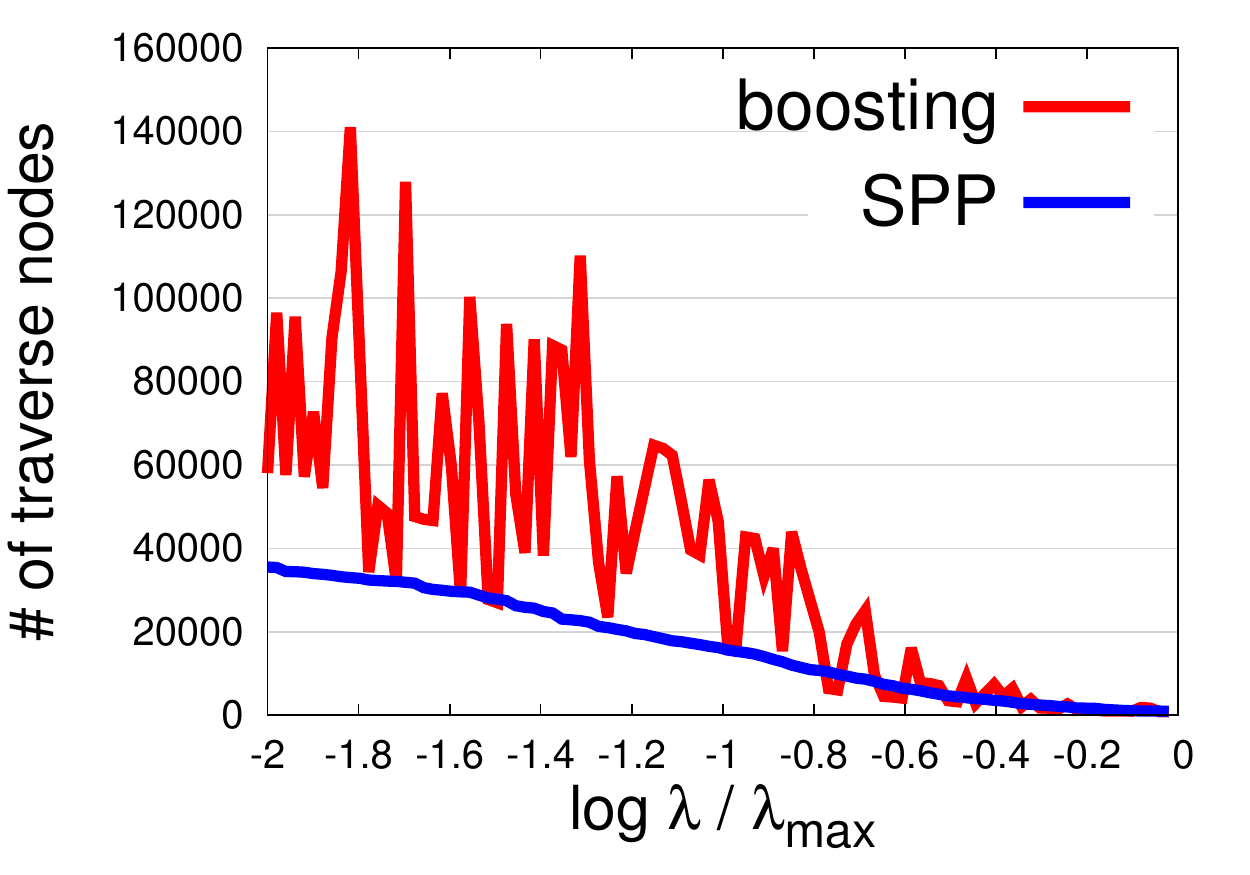} \\
(c-1) maxpat 6 &
(c-2) maxpat 7 &
(c-3) maxpat 8 &
(c-4) maxpat 10 \\\\
\end{tabular}}
\subfigure[{\tt Karthikeyan}]{
\begin{tabular}{cccc}
\includegraphics[scale=0.33]{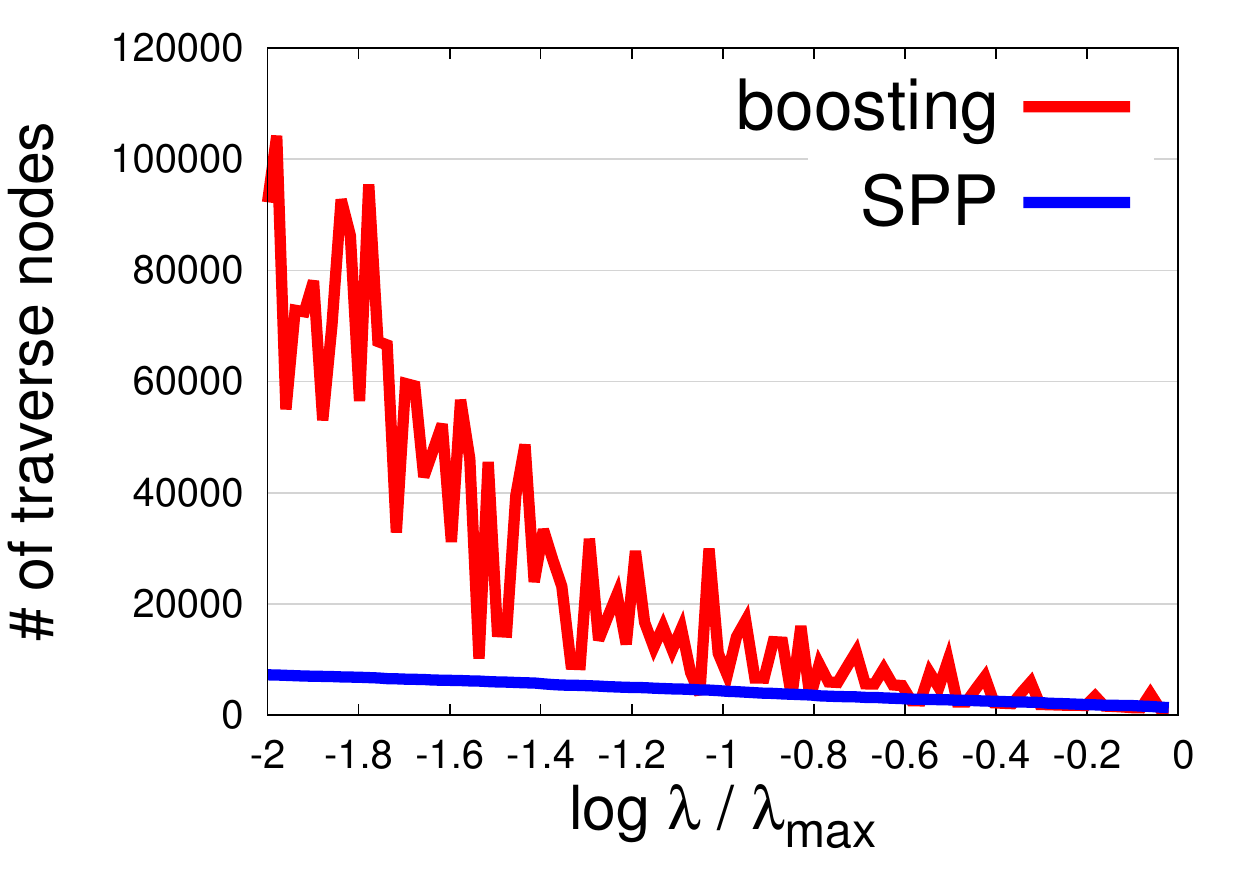} &
\includegraphics[scale=0.33]{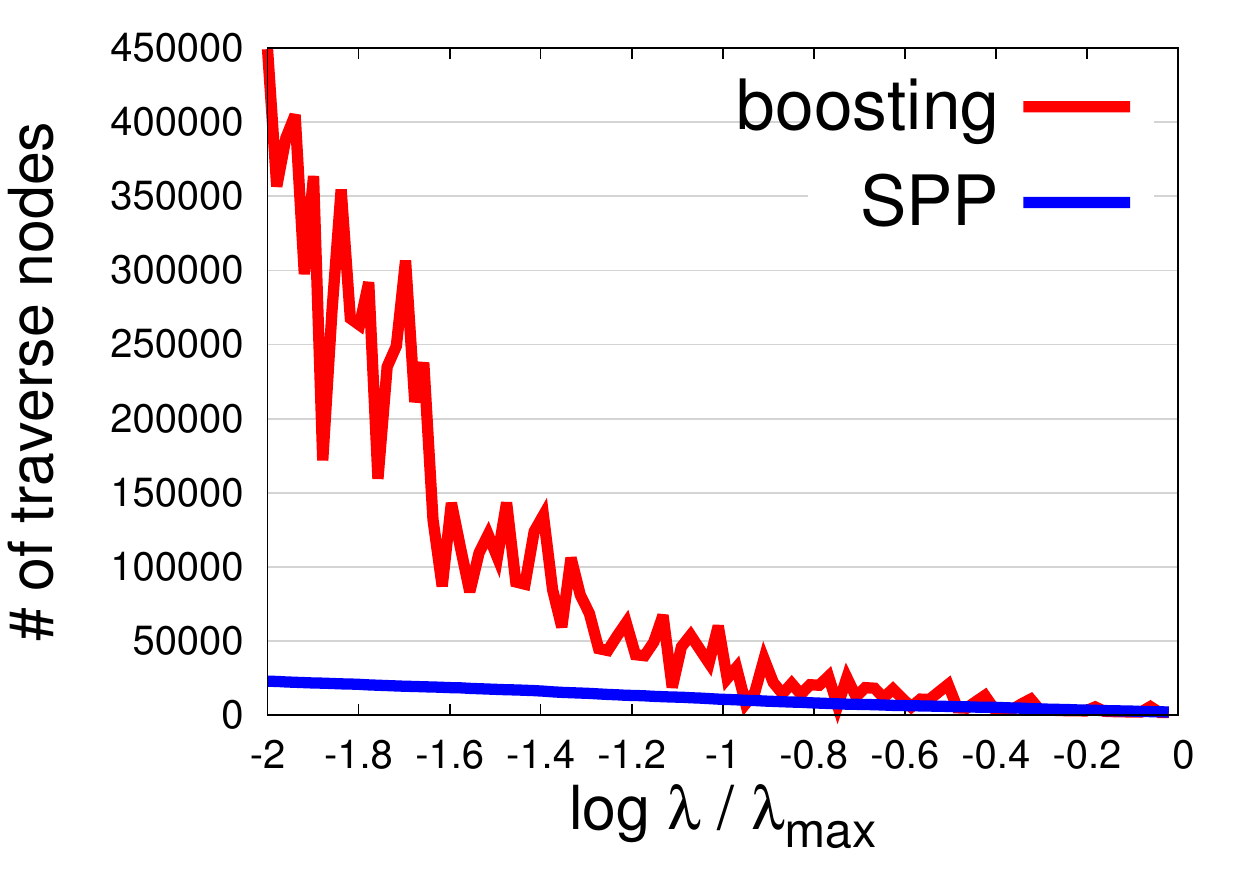} &
\includegraphics[scale=0.33]{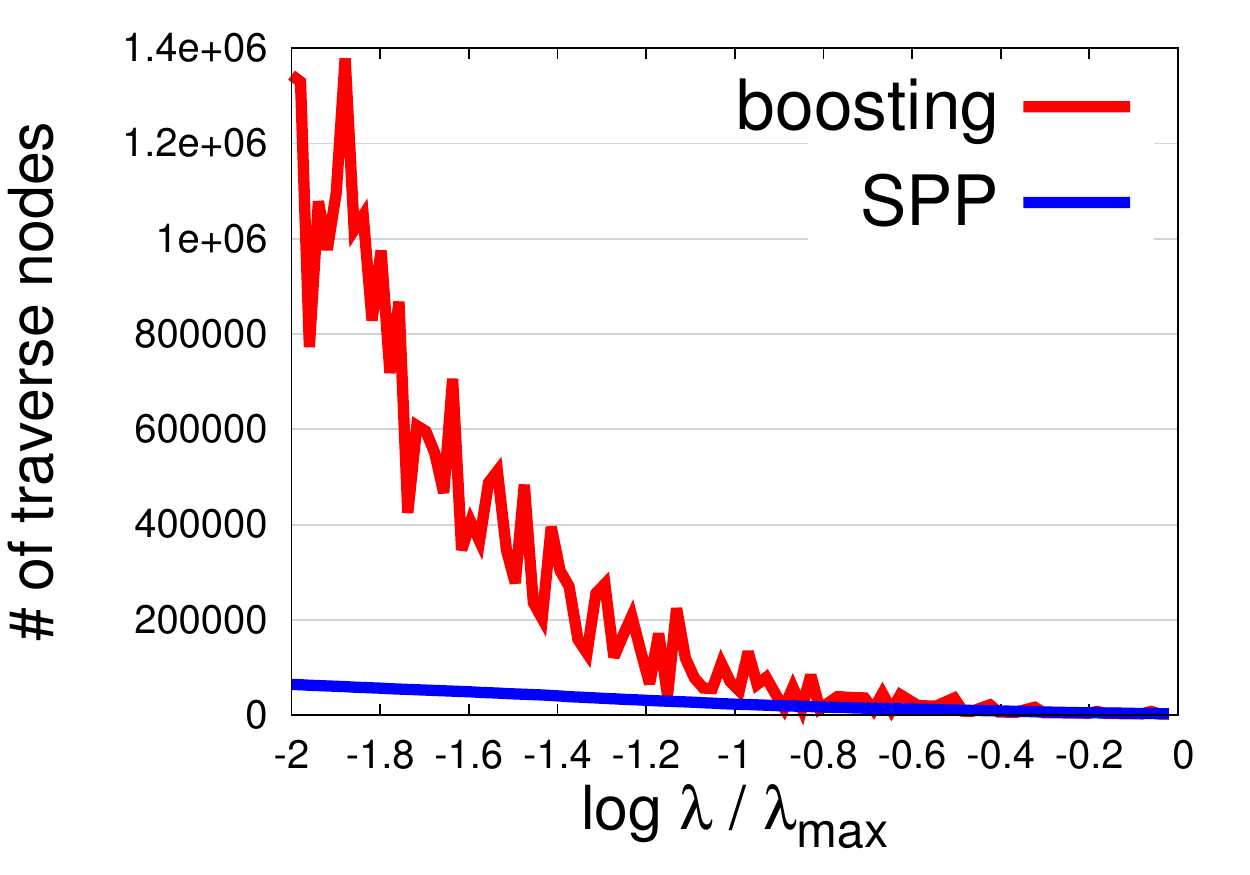} &
\includegraphics[scale=0.33]{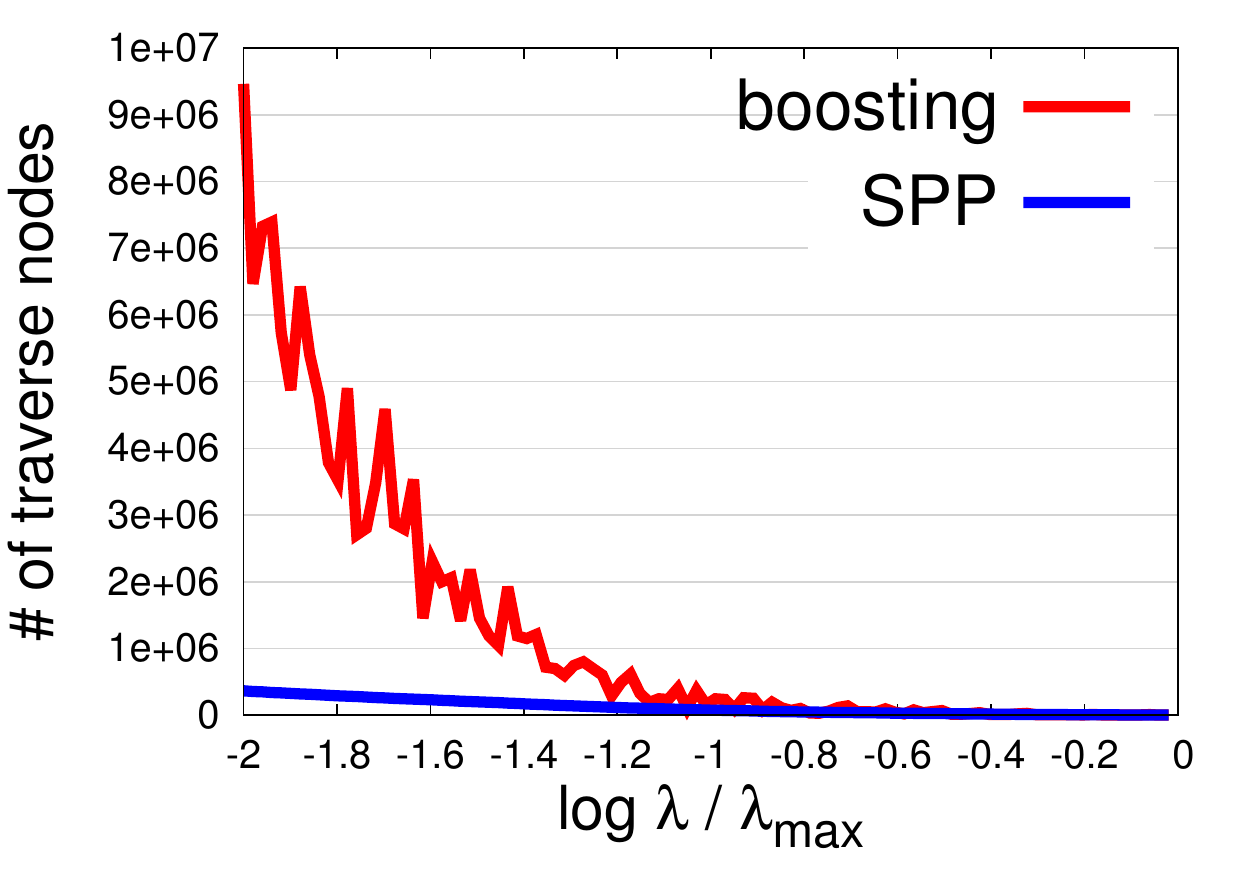} \\
(d-1) maxpat 6 &
(d-2) maxpat 7 &
(d-3) maxpat 8 &
(d-4) maxpat 10 \\\\
\end{tabular}}
\end{center}
\caption{\# of traverse nodes for graph classification and regression.}
\label{fig:graph_traverse_c}
\end{figure}

\figurename~\ref{fig:graph_time_c} shows the computation time of the two methods.
In all the cases, {\tt SPP} is faster than {\tt boosting}, and the difference gets larger as {\tt maxpat} increases.  
\figurename~\ref{fig:graph_time_c} also shows the computation time taken in traversing the trees ({\tt traverse}) and that taken in solving the optimization problems ({\tt solve}).
The results indicate that {\tt traverse} time of {\tt SPP} are only slightly better than that of {\tt boosting}. 
It is because the most time-consuming component of gSpan is the
minimality check of the DFS (depth-first search) code, and the {\tt traverse} time mainly depends on how many different nodes are generated in the entire regularization path computation process\footnote{A common trick used in graph mining algorithms with gSpan is to keep the minimality check results in the memory for all the nodes generated so far.}. 
In terms of {\tt solve} time, there are large difference between {\tt SPP} and {\tt boosting}. 
In {\tt SPP}, we have only to solve a single convex optimization problem for each $\lambda$.
In {\tt boosting}, on the other hand, convex optimization problems must be repeatedly solved every time a new pattern is added to the working set. 
%
%
\figurename~\ref{fig:graph_traverse_c} shows the total number of traversed nodes in the entire regularization path computation process. 
%
Total number of traversed nodes in {\tt SPP} is much smaller than those of {\tt boosting}, which is because one must repeat searching over trees many times in {\tt boosting}. 
%

\subsection{Item-set classification/regression}

We applied {\tt SPP} and {\tt boosting} to item-set classification and regression problems. 
For classification, we used 
{\tt splice} dataset ($n = 1000$ and the
number of items $d = 120$) and
{\tt a9a} dataset ($n = 32561$ and $d = 123$).
For regression, we used {\tt dna} dataset ($n = 2000$ and $d = 180$) 
and {\tt protein} dataset ($n = 6621$ and $d = 714$)\footnote{This dataset is provided for
classification. We used it for regression simply by regarding the class
label as the scalar response variable.}.
%
%
All datasets were obtained from LIBSVM Dataset site \cite{CC01a}.
We considered the cases with {\tt maxpat} $\in \{3, 4, 5, 6\}$, where {\tt maxpat} here indicates the maximum size of item-sets we wanted to find.

\begin{figure}[t]
\begin{center}
\subfigure[{\tt splice}]{
\includegraphics[clip,width=0.21\textwidth]{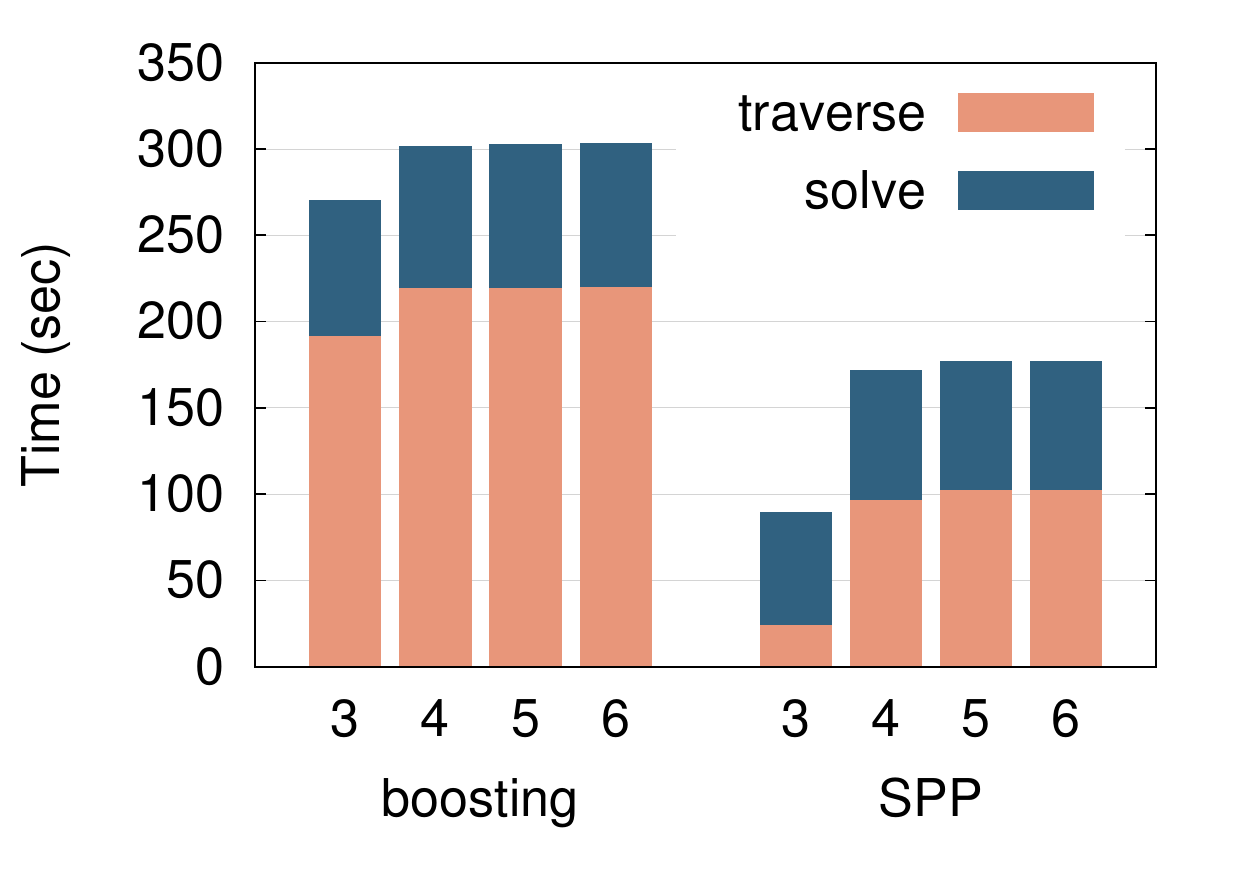}}
\subfigure[{\tt a9a}]{
\includegraphics[clip,width=0.21\textwidth]{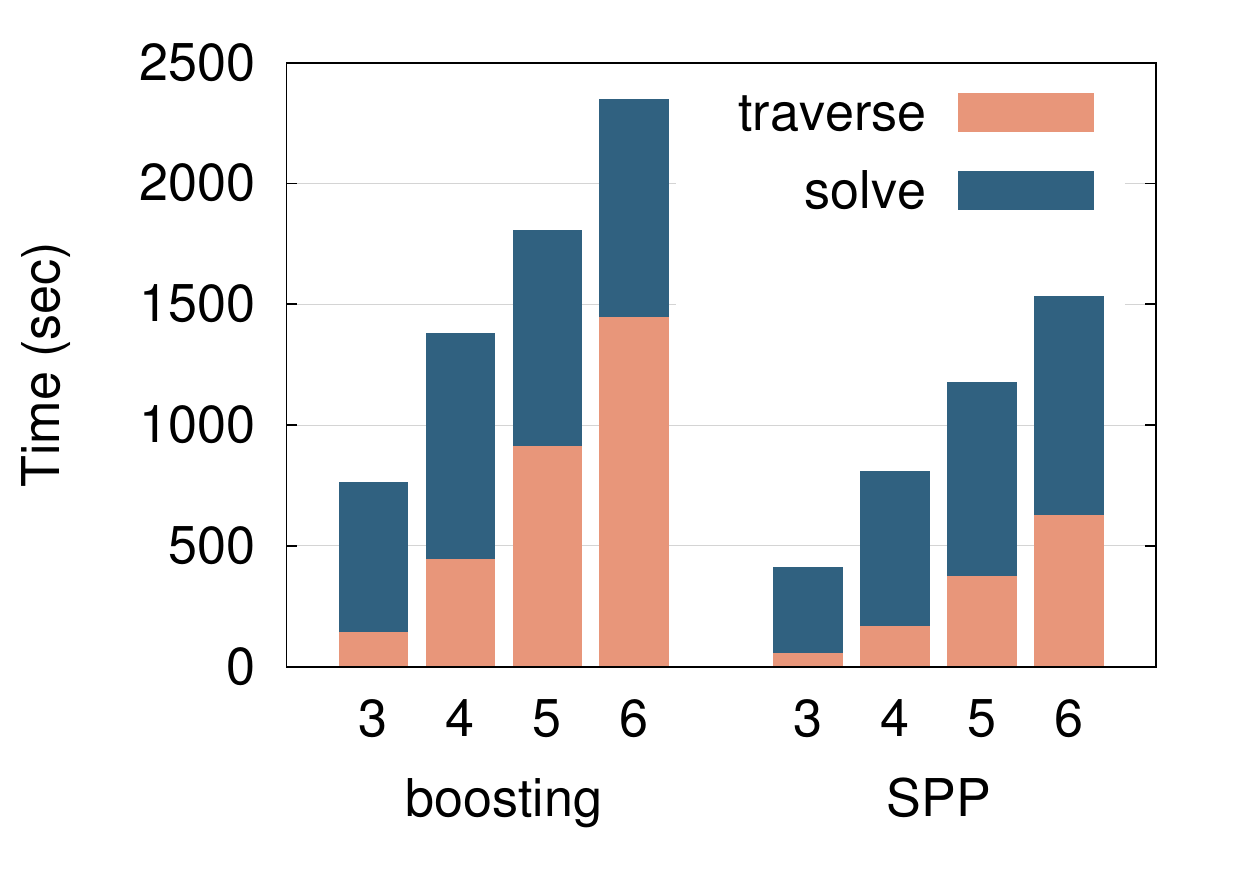}}
\subfigure[{\tt dna}]{
\includegraphics[clip,width=0.21\textwidth]{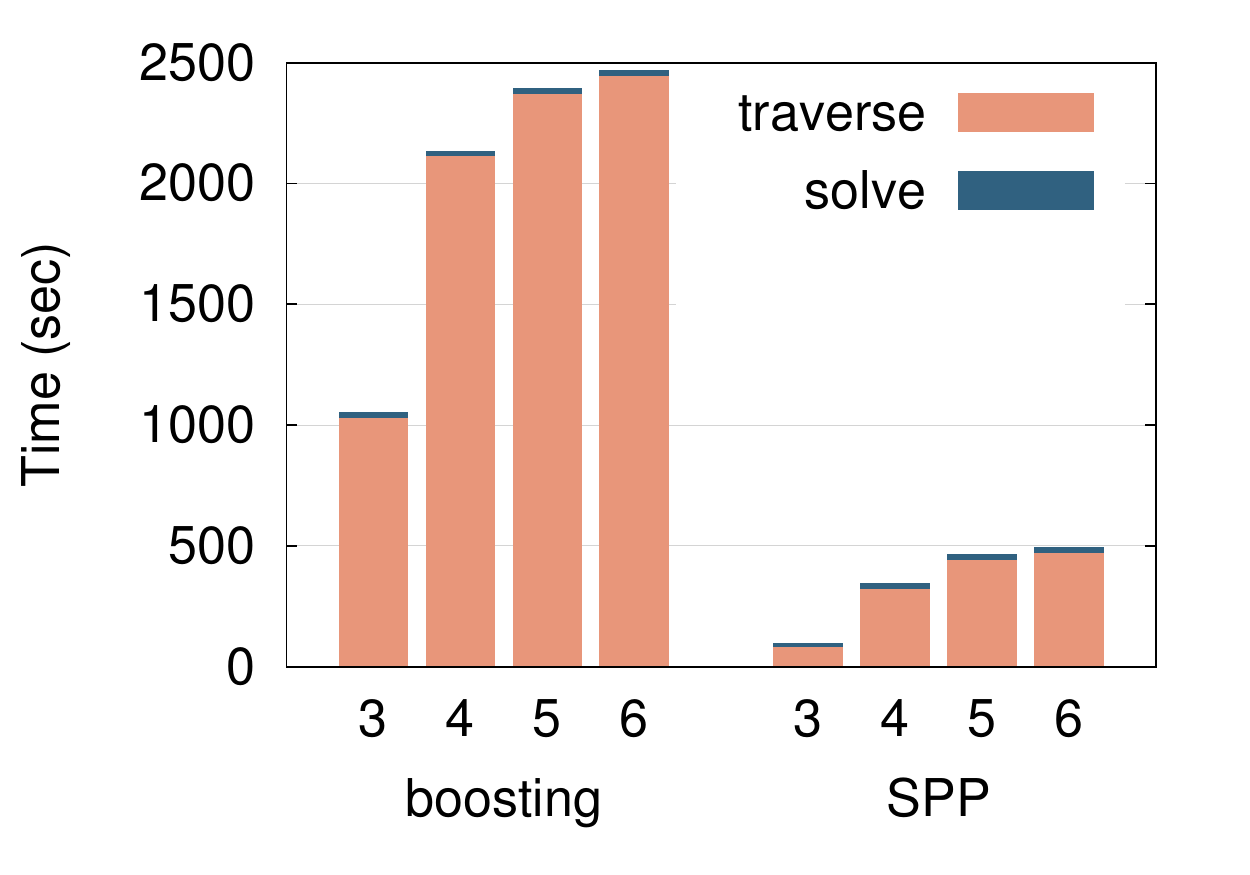}} 
\subfigure[{\tt protein}]{
\includegraphics[clip,width=0.21\textwidth]{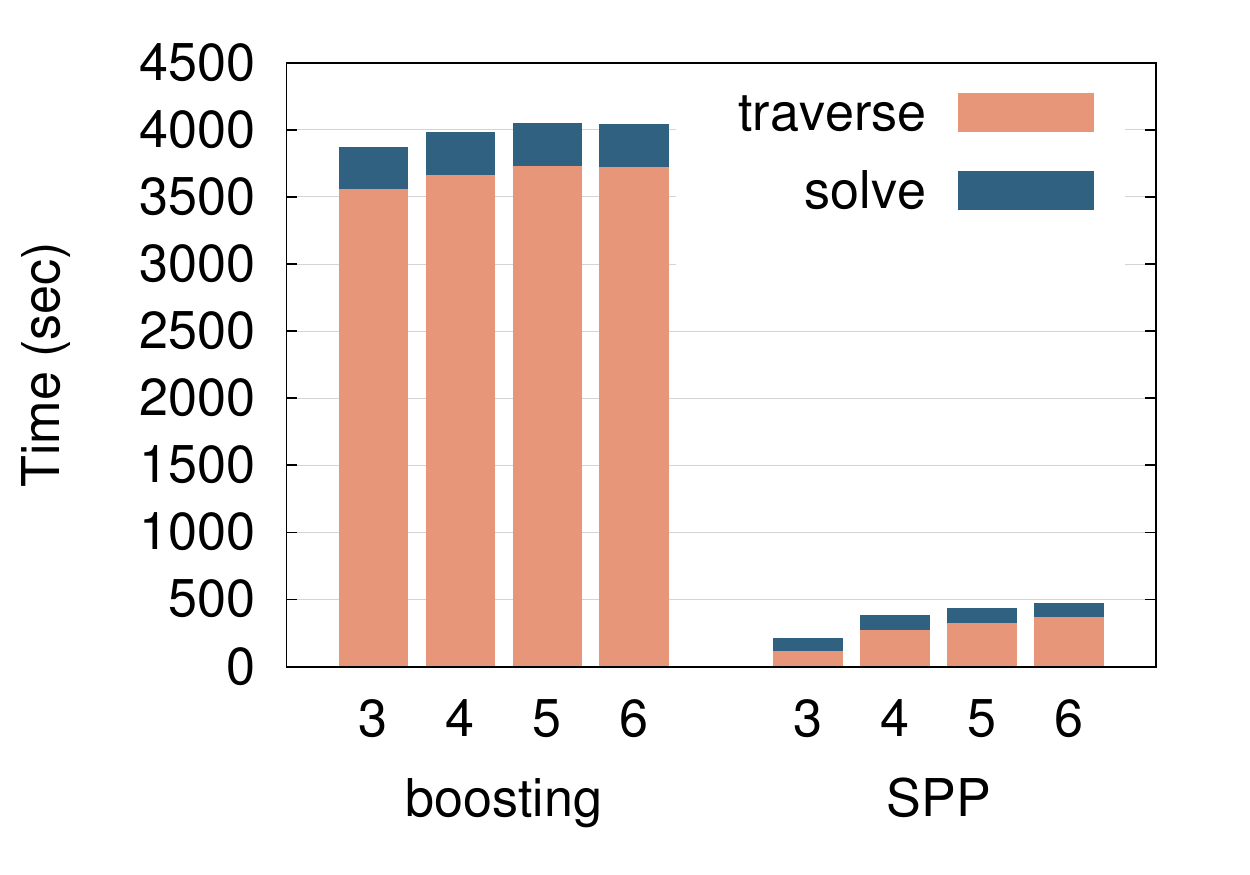}}
\caption{Computation time comparison for item-set classification and
 regression. Each bar contains computational time taken in the tree
 traverse ({\tt traverse}) and the optimization procedure ({\tt solve}) respectively.}
\label{fig:item_time_c}
\end{center}
\end{figure}

\begin{figure}
\begin{center}
\subfigure[{\tt splice}]{
\begin{tabular}{cccc}
\includegraphics[scale=0.33]{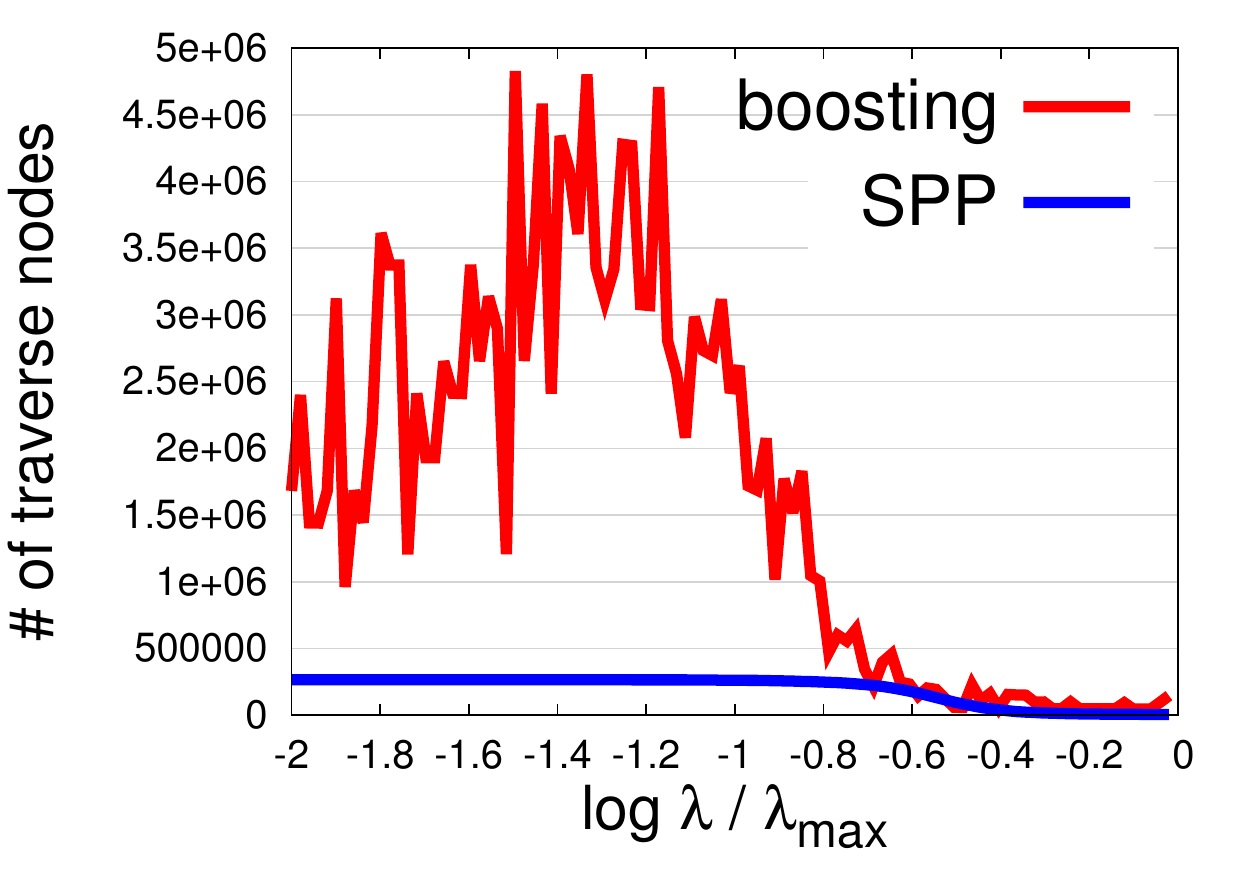} &
\includegraphics[scale=0.33]{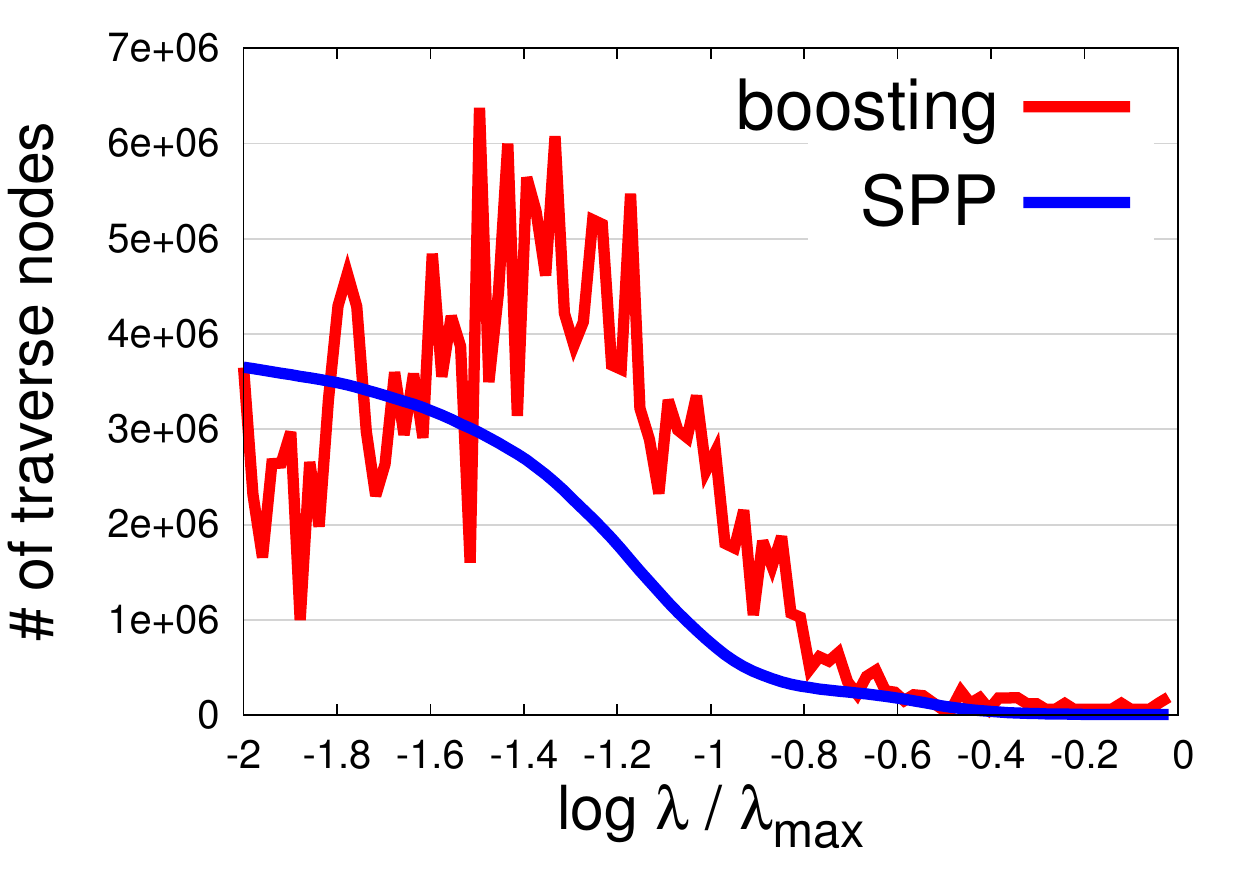} &
\includegraphics[scale=0.33]{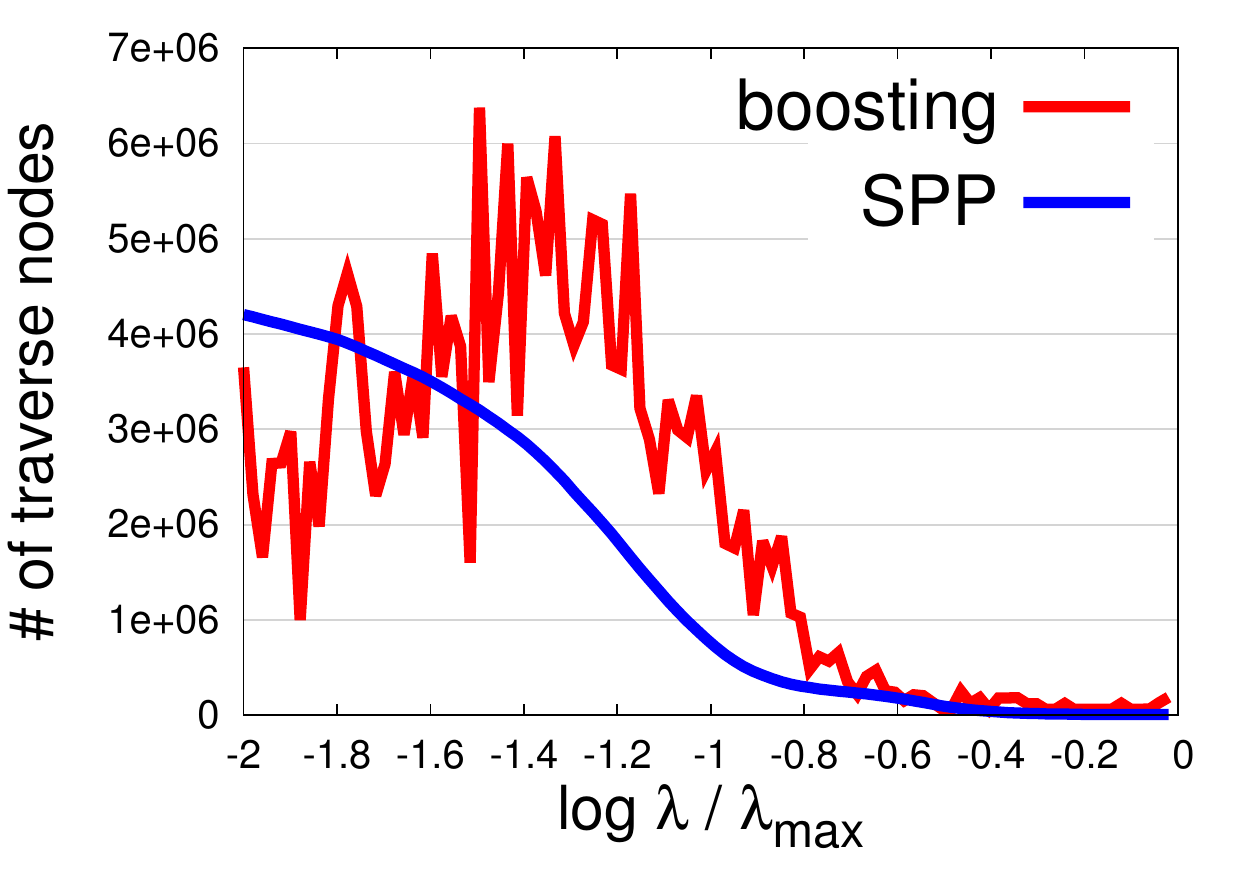} &
\includegraphics[scale=0.33]{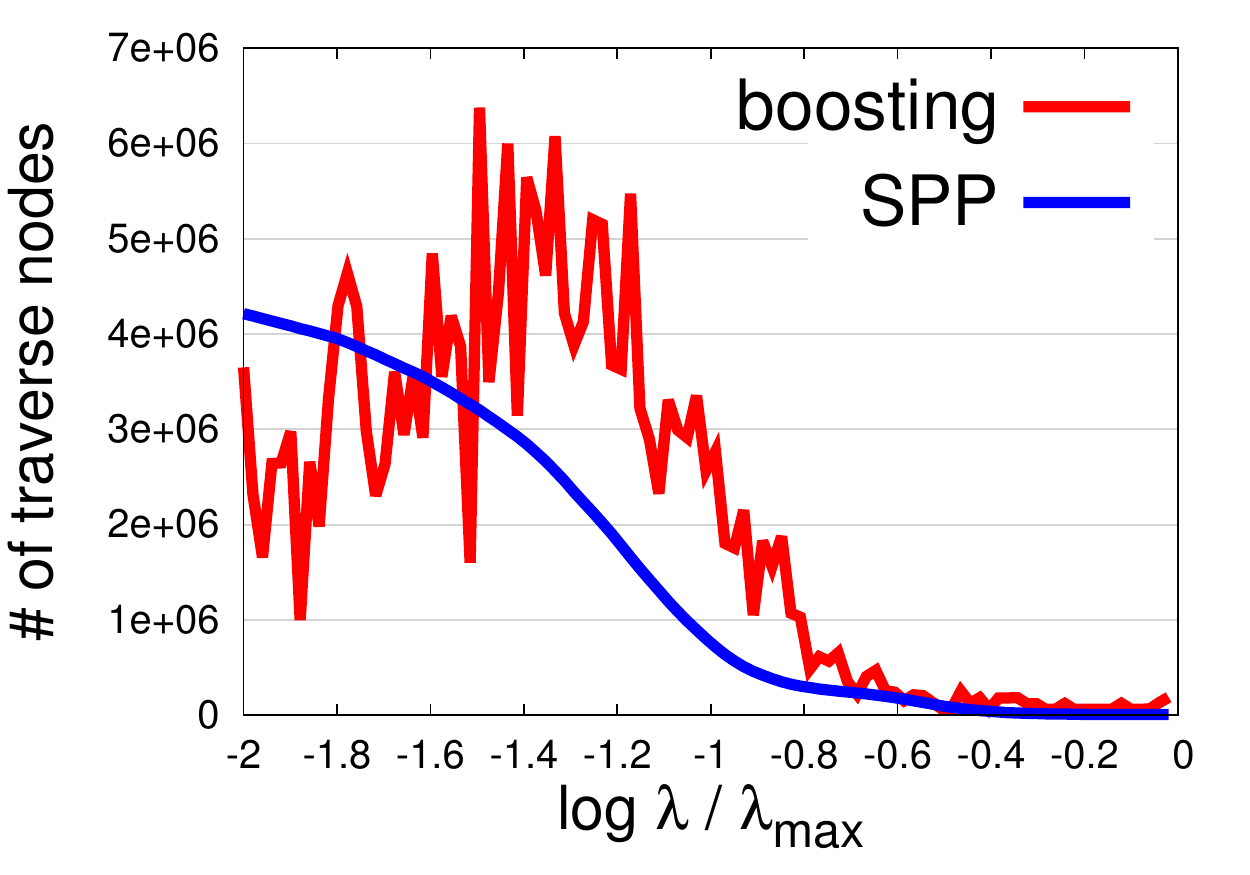} \\
(a-1) maxpat 3 &
(a-2) maxpat 4 &
(a-3) maxpat 5 &
(a-4) maxpat 6 \\\\
\end{tabular}
}
\subfigure[{\tt a9a}]{
\begin{tabular}{cccc}
\includegraphics[scale=0.33]{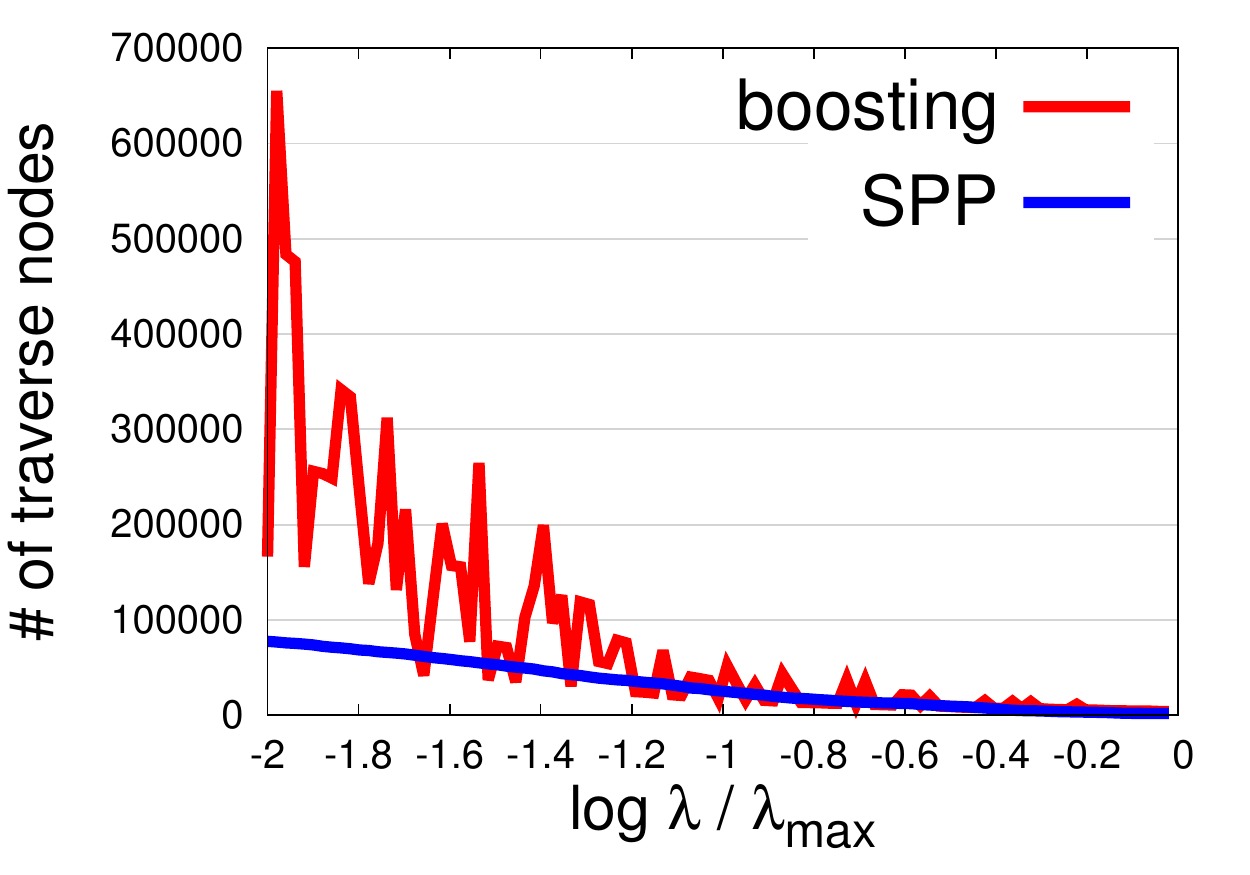} &
\includegraphics[scale=0.33]{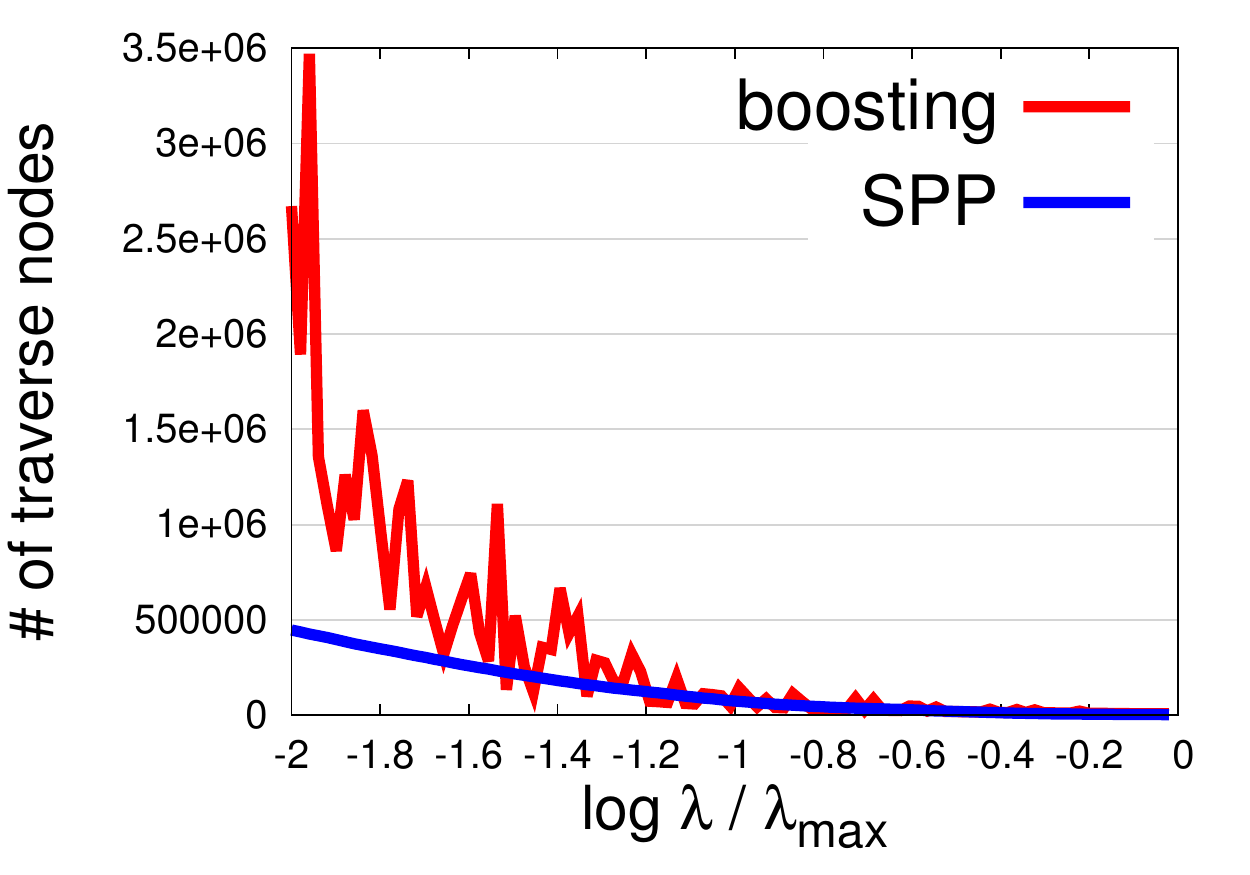} &
\includegraphics[scale=0.33]{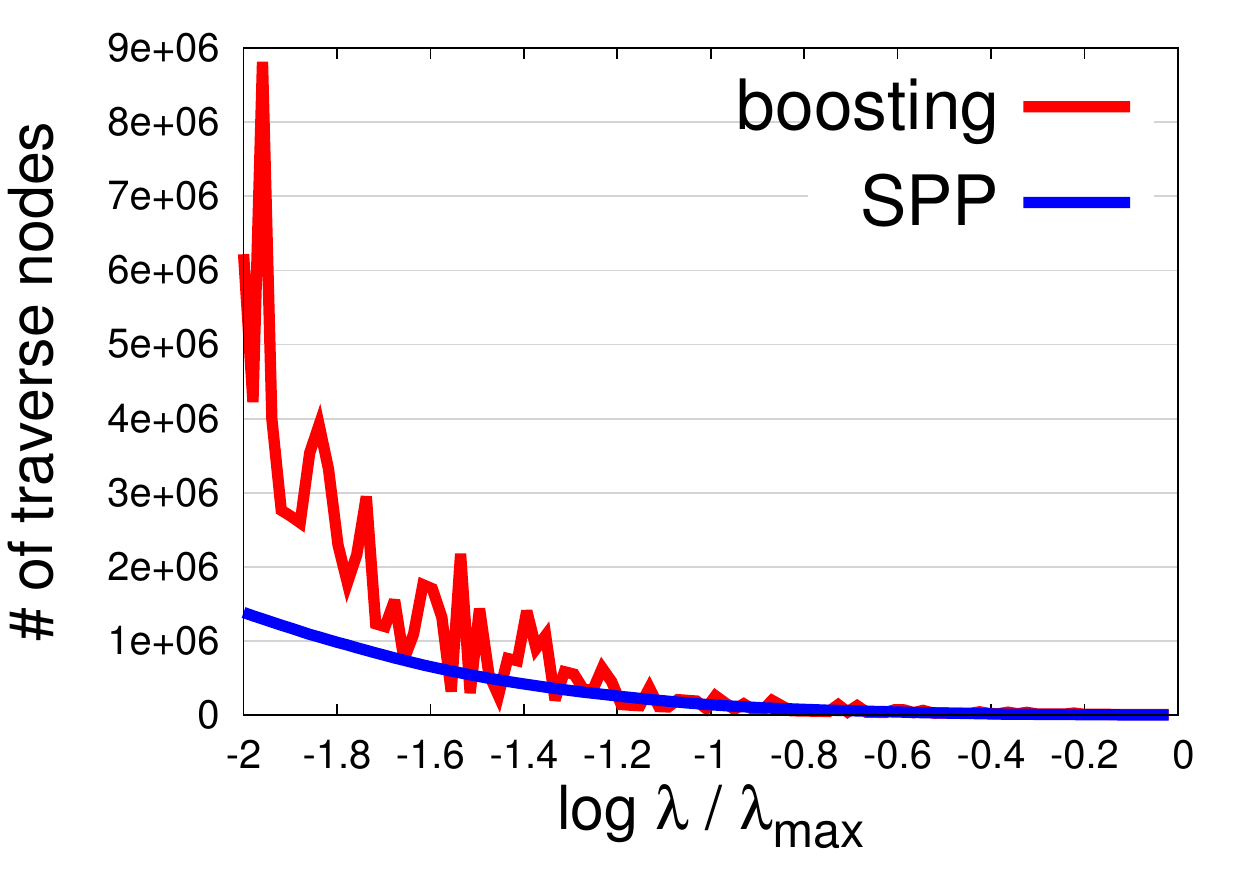} &
\includegraphics[scale=0.33]{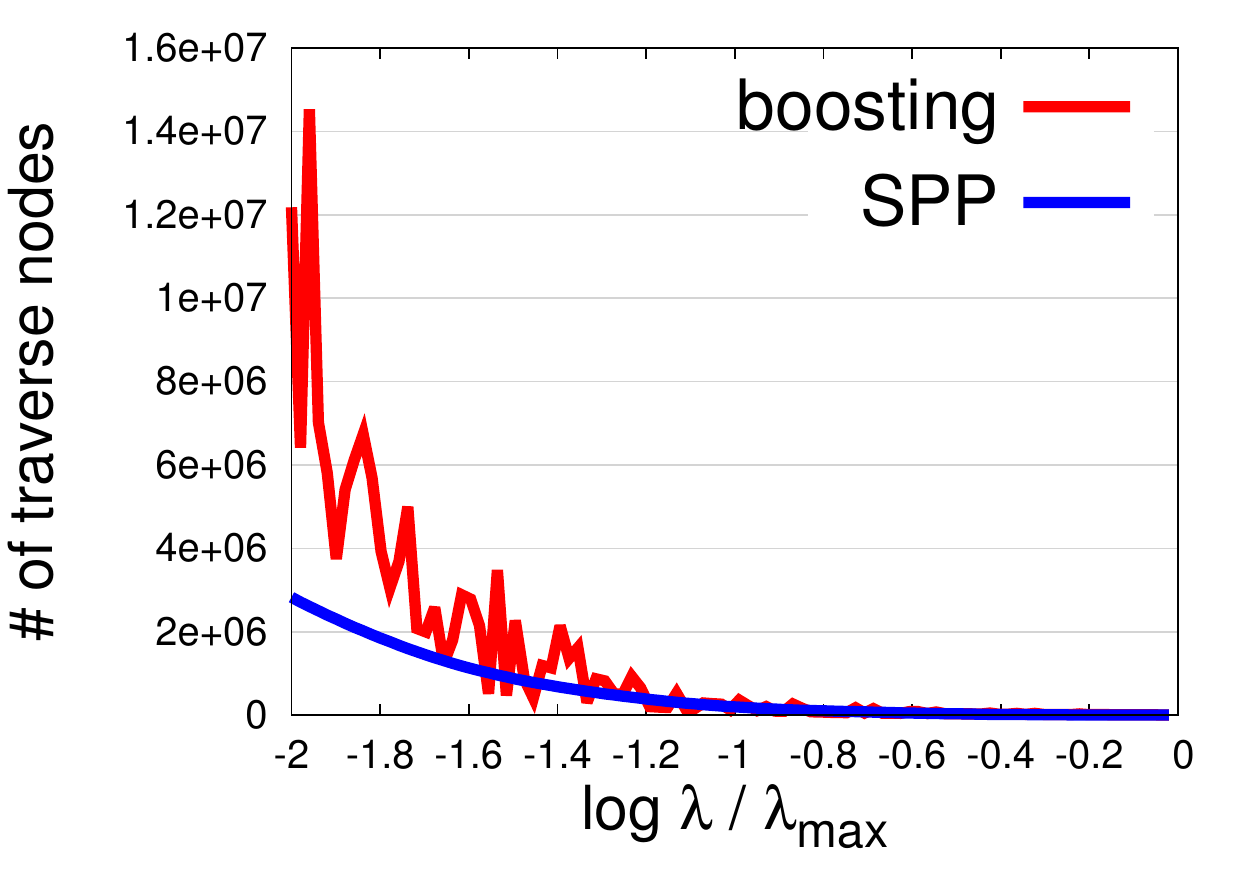} \\
(b-1) maxpat 3 &
(b-2) maxpat 4 &
(b-2) maxpat 5 &
(b-3) maxpat 6 \\\\
\end{tabular}
}
\subfigure[{\tt dna}]{
\begin{tabular}{cccc}
\includegraphics[scale=0.33]{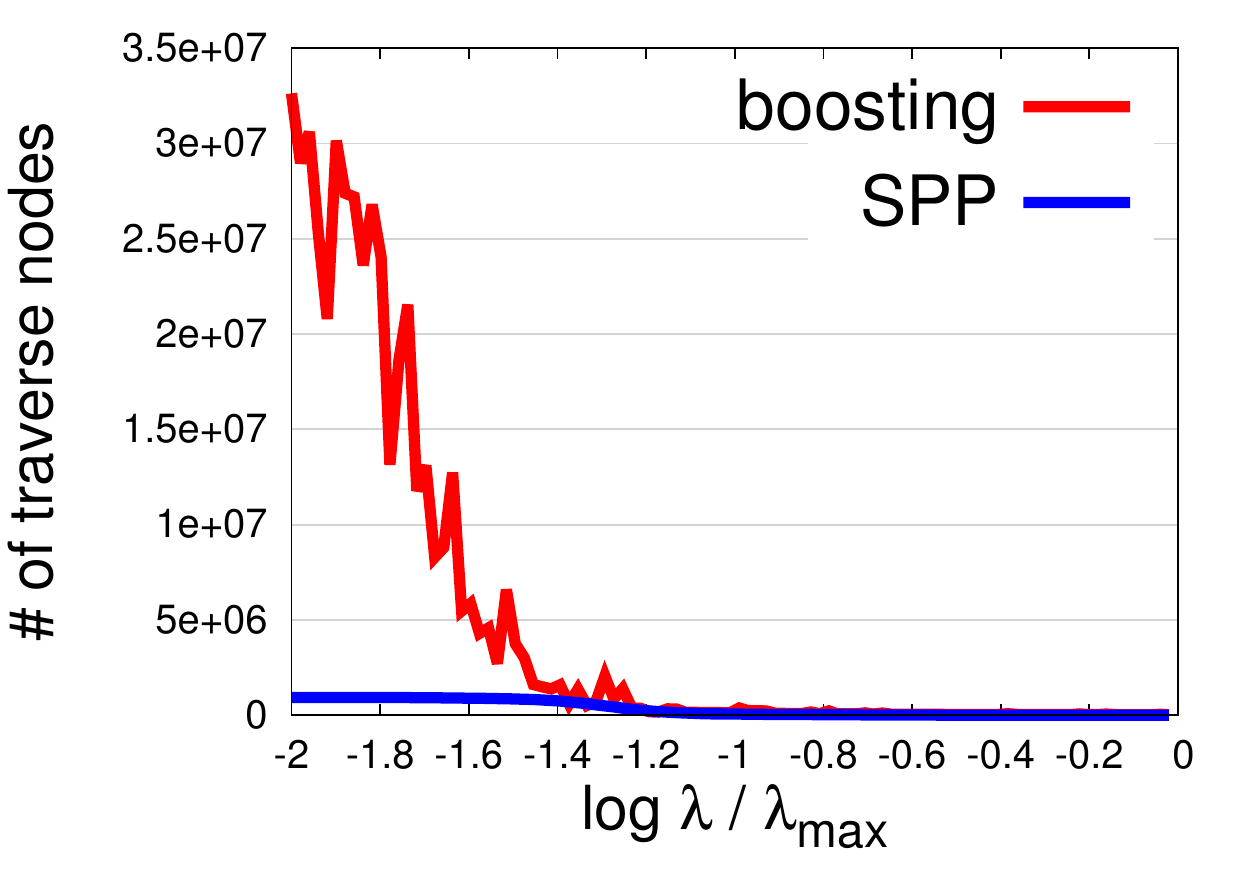} &
\includegraphics[scale=0.33]{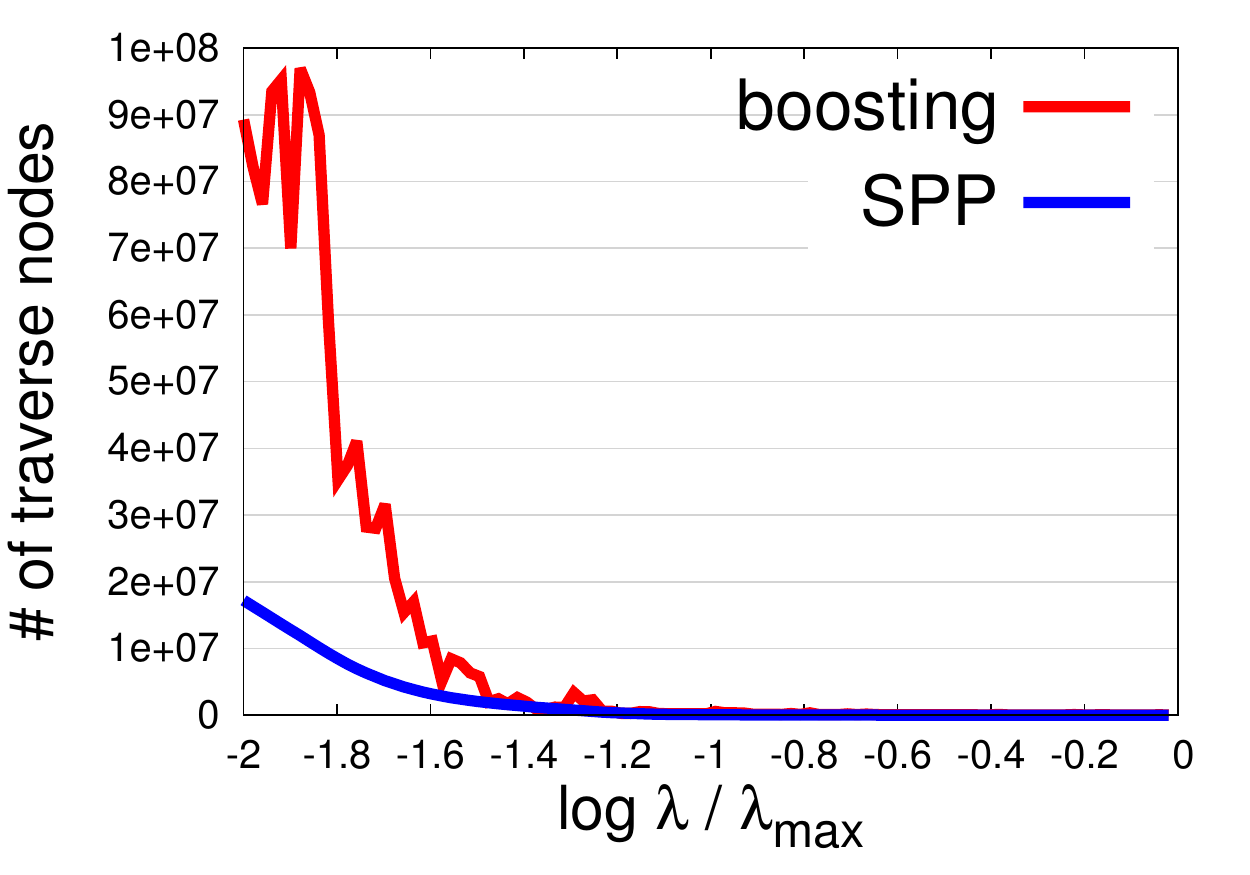} &
\includegraphics[scale=0.33]{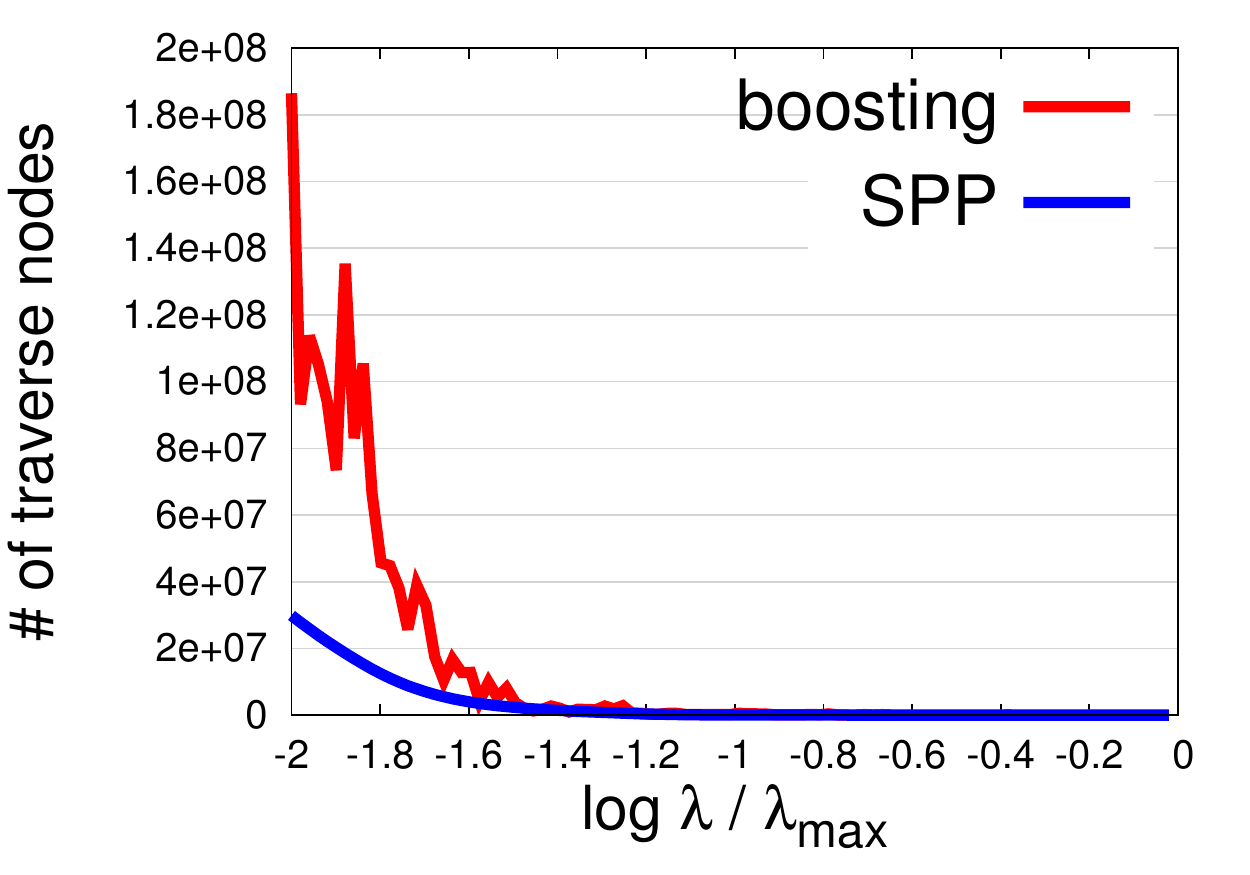} &
\includegraphics[scale=0.33]{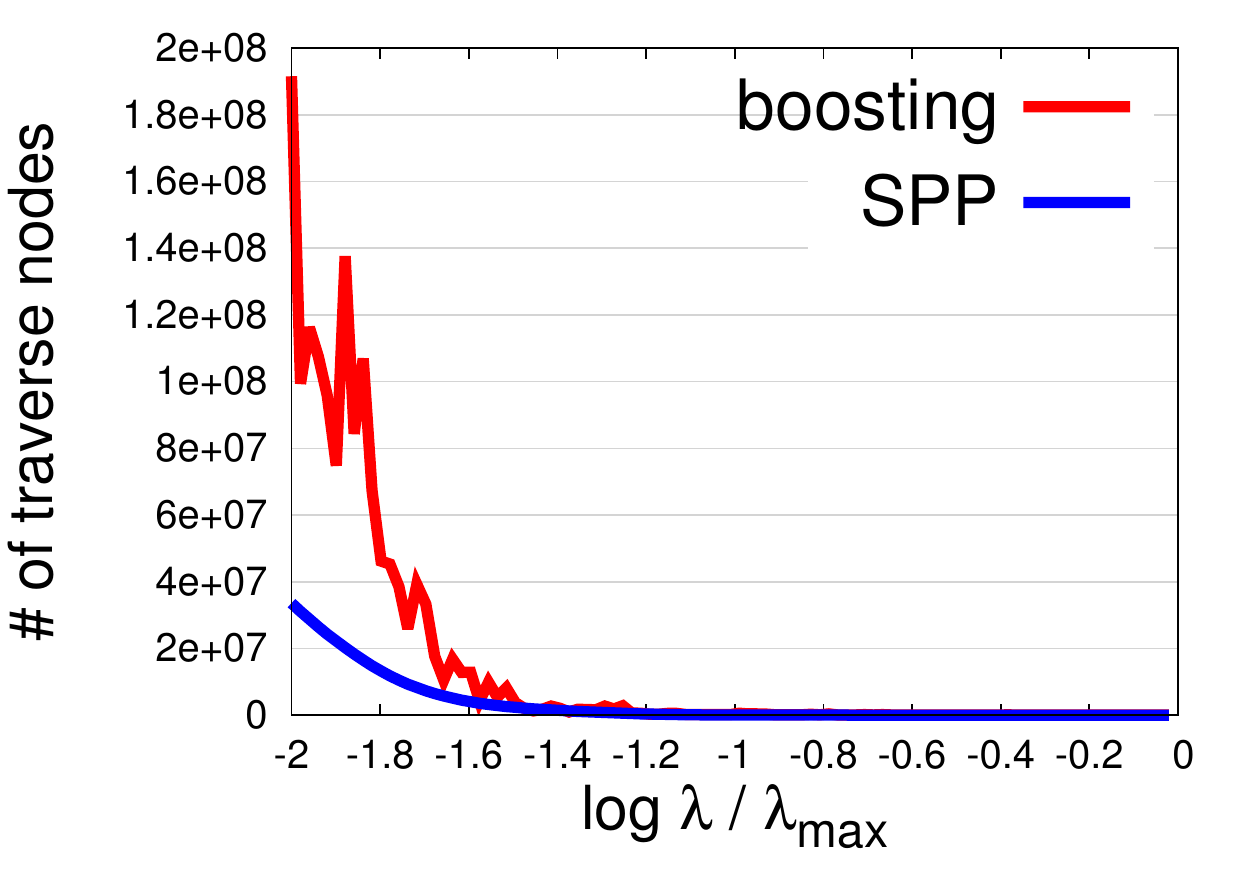} \\
(c-1) maxpat 3 &
(c-2) maxpat 4 &
(c-3) maxpat 5 &
(c-4) maxpat 6 \\\\
\end{tabular}
}
\subfigure[{\tt protein}]{
\begin{tabular}{cccc}
\includegraphics[scale=0.33]{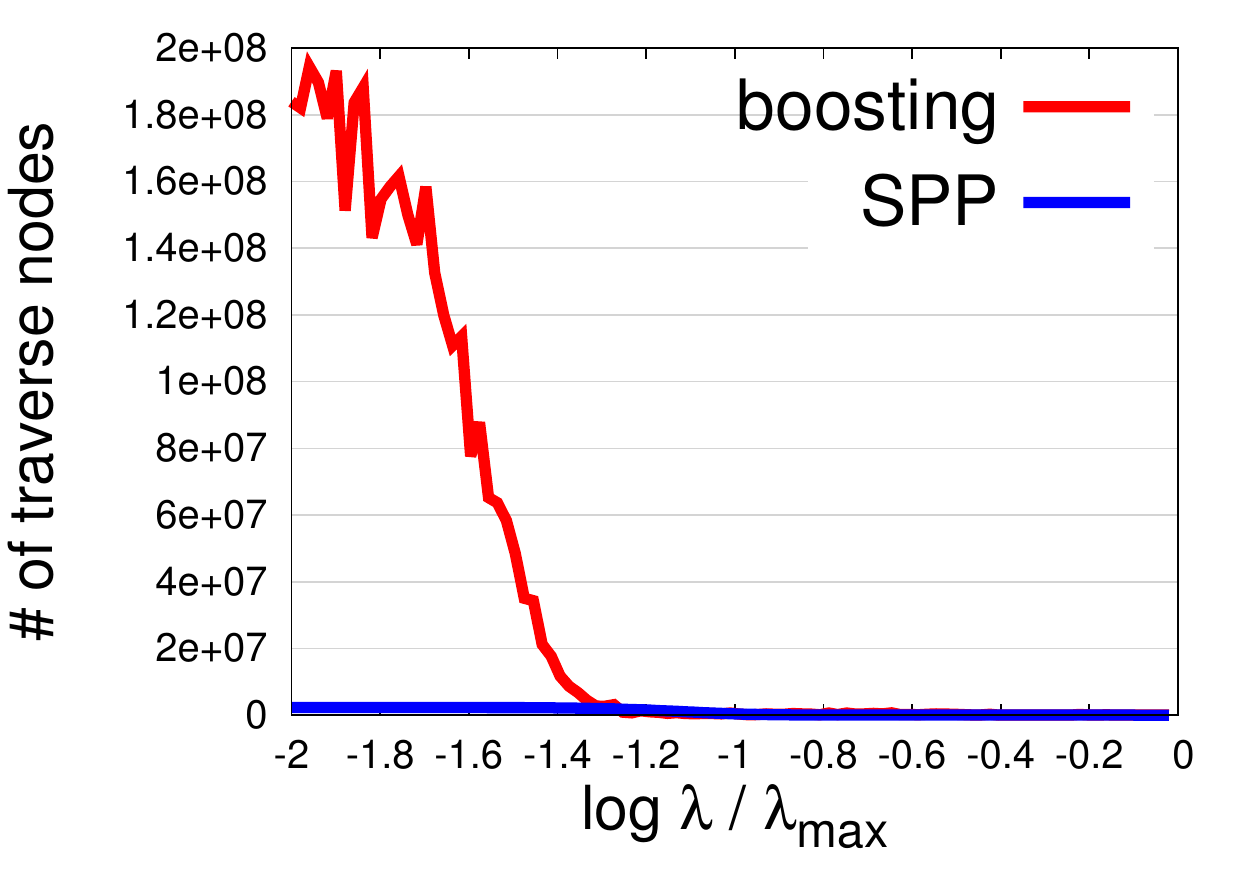} &
\includegraphics[scale=0.33]{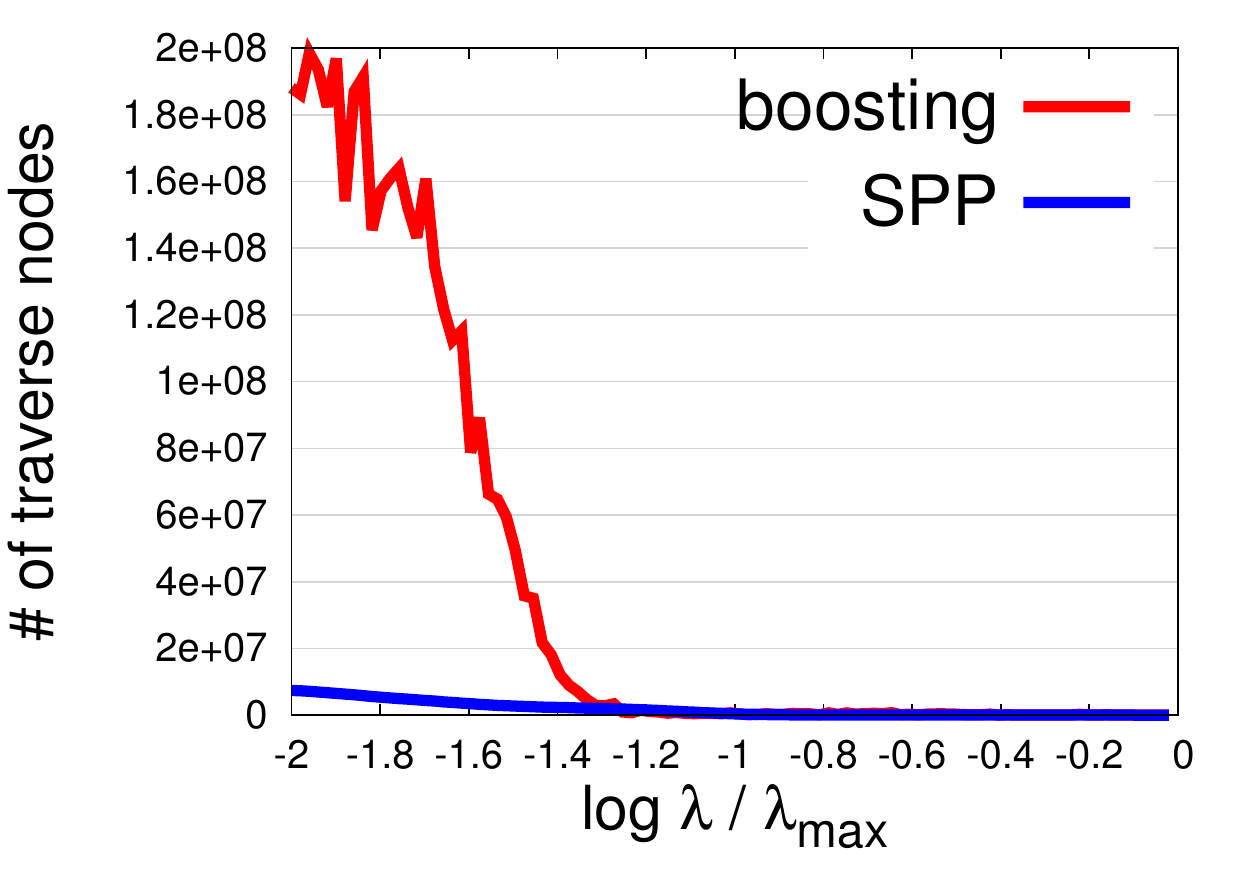} &
\includegraphics[scale=0.33]{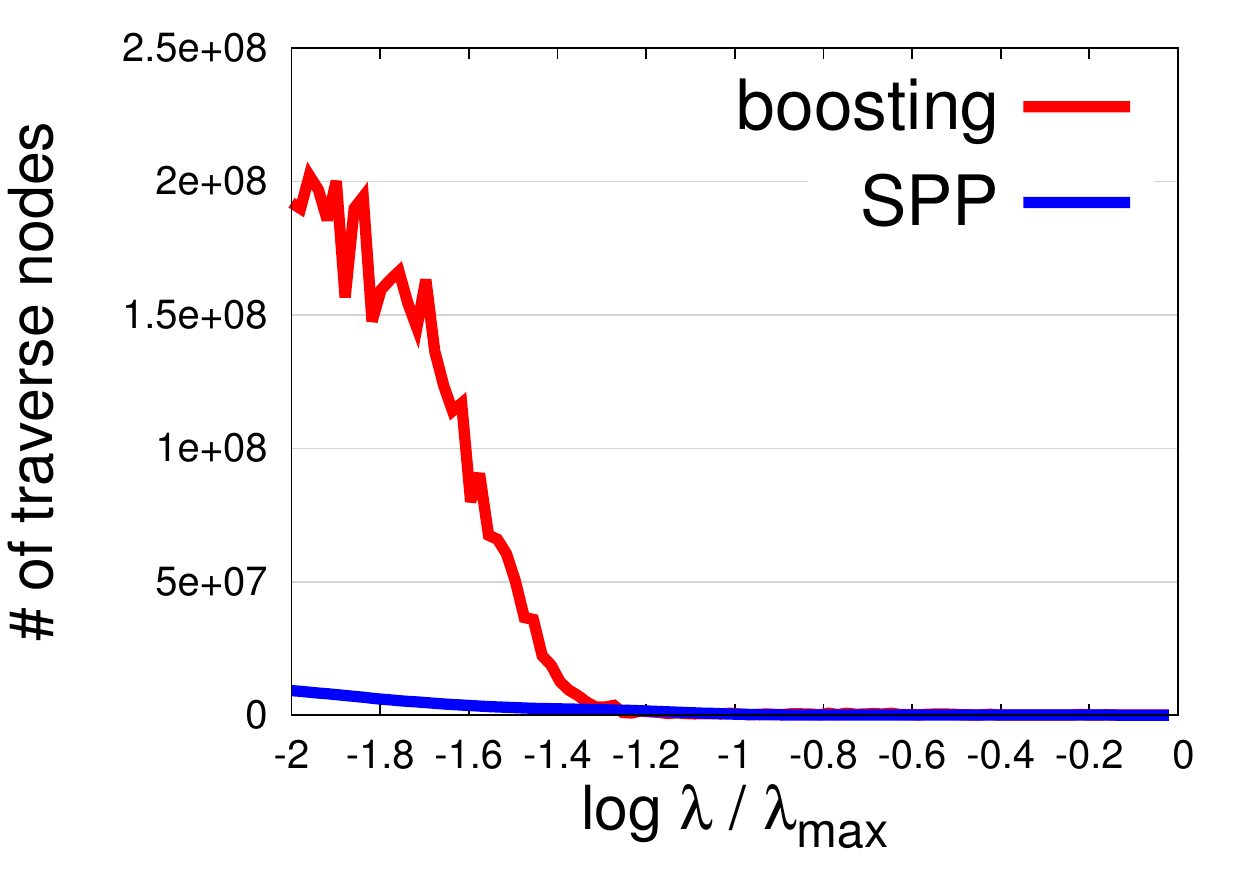} &
\includegraphics[scale=0.33]{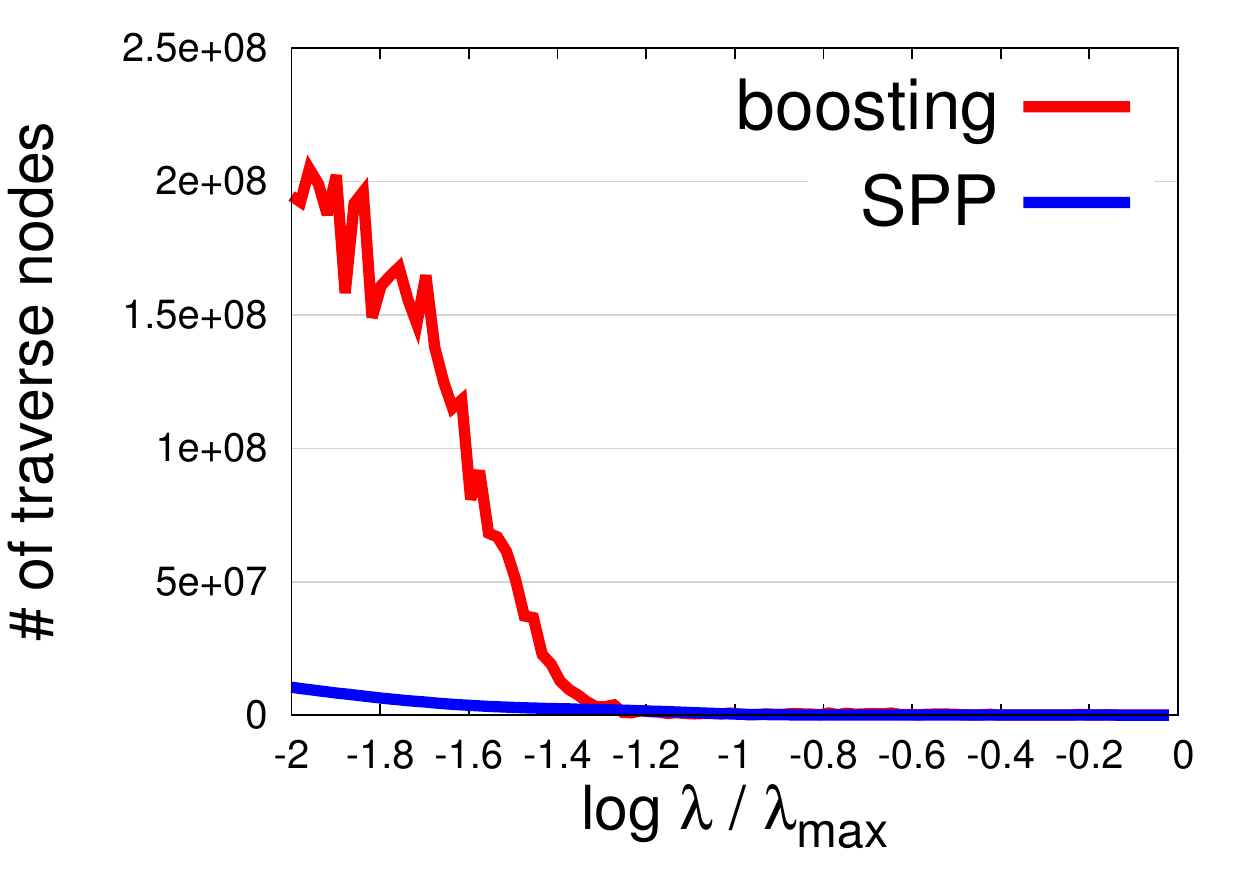} \\
(d-1) maxpat 3 &
(d-2) maxpat 4 &
(d-3) maxpat 5 &
(d-4) maxpat 6 \\\\
\end{tabular}
}
\end{center}
\caption{\# of traverse nodes for item-set classification and regression.}
\label{fig:item_traverse_c}
\end{figure}

\figurename~\ref{fig:item_time_c} compares the computation time of the two methods.
%
In all the cases, {\tt SPP} is faster than {\tt boosting}. 
Here again, 
\figurename~\ref{fig:item_time_c} also shows the computation time taken in traversing the trees ({\tt traverse}) and that taken in solving the optimization problems ({\tt solve}).
In contrast to the graph mining results, {\tt traverse} time of {\tt SPP} are much smaller than that of {\tt boosting} because it simply depends on how many nodes are traversed in total. 
\figurename~\ref{fig:item_traverse_c} shows the total number of traversed nodes in the entire regularization path computation process. 
Especially when $\lambda$ is small where the number of active patterns are large, {\tt boosting} needed to traverse large number of nodes,
which is because the number of steps of {\tt boosting} is large when there are large number of active patterns. 

%% file: Sec/sec5.tex
\section{Conclusions}
\label{sec:conclusions}
In this paper, we introduced a novel method called safe pattern pruning (SPP) method for a class of predictive pattern mining problems.
The advantage of the SPP method is that it allows us to efficiently find a superset of all the predictive patterns that are used in the optimal predictive model by a single search over the database.
We demonstrated the computational advantage of the SPP method by applying it to graph classification/regression and item-set classification/regression problems.
As a future work, we will study how to integrate the SPP method with a technique for providing the statistical significances of the discovered patterns \cite{suzumuraKDD2016submitted, terada2013statistical}.

%% file: App/appA.tex
\section{Proofs}

\paragraph{Proof of Corollary~\ref{coro:getting-tighter}}
\begin{proof}
For any pair of a node $t$ and $t^\prime \in \cT_{\rm sub}(t)$, 
\begin{align}
\label{bound_plus}
\sum_{i: \beta_i \tilde{\theta}_i > 0} \alpha_{it} \tilde{\theta}_i \ge
\sum_{i: \beta_i \tilde{\theta}_i > 0} \alpha_{it^\prime} \tilde{\theta}_i, \\
\label{bound_minus}
\sum_{i: \beta_i \tilde{\theta}_i < 0} \alpha_{it} \tilde{\theta}_i \le
\sum_{i: \beta_i \tilde{\theta}_i < 0} \alpha_{it^\prime} \tilde{\theta}_i. 
\end{align}
First consider the case where 
$u_t = \sum_{i: \beta_i \tilde{\theta}_i > 0} \alpha_{it} \tilde{\theta}_i$. 
When
$u_{t^\prime} = \sum_{i: \beta_i \tilde{\theta}_i > 0} \alpha_{it^\prime} \tilde{\theta}_i$,
from \eq{bound_plus}, 
$u_t \ge u_{t^\prime}$.
When 
$u_{t^\prime} = -\sum_{i: \beta_i \tilde{\theta}_i < 0} \alpha_{it^\prime} \tilde{\theta}_i$,
from \eq{bound_minus},
\begin{align*}
u_t \ge -\sum_{i: \beta_i \tilde{\theta}_i < 0} \alpha_{it} \tilde{\theta}_i  \ge u_{t^\prime}. 
\end{align*}
Next, consider the case where
$u_t = -\sum_{i: \beta_i \tilde{\theta}_i < 0} \alpha_{it} \tilde{\theta}_i$. 
When
$u_{t^\prime} = \sum_{i: \beta_i \tilde{\theta}_i > 0} \alpha_{it^\prime} \tilde{\theta}_i$,
from \eq{bound_plus},
\begin{align*}
 u_t \ge \sum_{i: \beta_i \tilde{\theta}_i < 0} \alpha_{it} \tilde{\theta}_i  \ge u_{t^\prime}. 
\end{align*}
When
$u_{t^\prime} = -\sum_{i: \beta_i \tilde{\theta}_i < 0} \alpha_{it^\prime} \tilde{\theta}_i$,
from \eq{bound_minus},
$u_t \ge u_{t^\prime}$. 
Furthermore,
it is clear that
$v_t \ge v_{t^\prime}$.
Since
$r_\lambda > 0$,
${\rm SPPC}(t) \ge {\rm SPPC}(t^\prime)$. 
\end{proof}

\paragraph{Proof of Lemma~\ref{lemm:1st}}
\begin{proof}
Based on the convex optimization theory (see, e.g., \cite{boyd2004convex}),
the KKT optimality condition of the primal problem
\eq{eq:general-problem}
and the dual problem 
\eq{eq:dual-problem}
is written as
\begin{align*}
\sum_{i=1}^n \alpha_{it} \theta_i^* \in
\begin{cases}
{\rm sign}(w_t^*)  & {\rm if~} w_t^* \ne 0, \\
[-1, +1] & {\rm if~} w_t^* = 0, 
\end{cases}
~~\forall t \in \cT. 
\end{align*}
It suggests that
\begin{align*}
\left|
\sum_{i=1}^n \alpha_{it} \theta_i^*
\right| < 1
~\Rightarrow~
w_t^* = 0,~
\forall t \in \cT. 
\end{align*}
\end{proof}

\paragraph{Proof of Lemma~\ref{lemm:3rd}}
\begin{proof}
 Let
 $\bm \alpha_{:,t} := [\alpha_{1t}, \ldots, \alpha_{nt}]^\top$. 
 First, 
 note that the objective part of the optimization problem
 \eq{eq:ub}
 is rewritten as
  \begin{align}
   \nonumber
   &
   \max_{\bm \theta}
   ~
   \left| \bm \alpha_{:,t}^\top \bm \theta \right|
   \\
   \nonumber
   ~\Leftrightarrow~
   &
   \max_{\bm \theta}
   \max \left\{
     \bm \alpha_{:,t}^\top \bm \theta,
   - \bm \alpha_{:,t}^\top \bm \theta
   \right\}
   \\
   \label{eq:lemm5-decompose}
   ~\Leftrightarrow~
   &
   \max 
   \left\{
   - \min_{\bm \theta} (- \bm \alpha_{:,t})^\top \bm \theta,
   - \min_{\bm \theta} \bm \alpha_{:,t}^\top \bm \theta
   \right\}
  \end{align}
 Thus,
 we consider the following convex optimization problem: 
 \begin{align}
  \label{subprob2}
   \min_{\bm \theta}~\bm \alpha_{:,t}^\top \bm \theta
   ~
   {\rm s.t.}~
   \|\bm \theta - \tilde{\bm \theta} \|_2^2 \le r_\lambda^2, 
   \bm \beta^\top \bm \theta = 0. 
 \end{align}
Let us define the Lagrange function
\begin{align*}
 L(\bm \theta, \xi, \eta) = 
 \bm \alpha_{:,t}^\top \bm \theta + 
 \xi (\| \bm \theta - \tilde{\bm \theta} \|^2_2 - r_\lambda^2) + \eta \bm \beta^\top \bm \theta, 
\end{align*}
and then the optimization problem 
\eq{subprob2}
is written as
\begin{align}
\label{minmax}
\min_{\bm \theta} \max_{\xi \ge 0, \eta} L(\bm \theta, \xi, \eta). 
\end{align}
The KKT optimality conditions are summarized as
\begin{subequations}
\begin{align}
\xi > 0, \\
\| \bm \theta - \tilde{\bm \theta} \|^2_2 - r_\lambda^2 \le 0, \\
\label{eq:KKT-eta}
\bm \beta^\top \bm \theta = 0, \\
\label{complement}
\xi (\| \bm \theta - \tilde{\bm \theta} \|^2_2 - r_\lambda^2) = 0, 
\end{align}
\end{subequations}
where note that $\xi > 0$ because the problem does not have a minimum value when $\xi = 0$. 
Differentiating the Lagrange function w.r.t. $\bm \theta$ and using the fact that it should be zero,
\begin{align}
\label{opt_theta}
\bm \theta = \tilde{\bm \theta} - \frac{1}{2\xi} (\bm \alpha_{:, t} + \eta \bm \beta).
\end{align}
By substituting
\eq{opt_theta}
into 
\eq{minmax},
\begin{align*}
\max_{\xi > 0, \eta} 
-\frac{1}{4 \xi} \| \bm \alpha_{:,t} + \eta \bm \beta \|^2_2 +
(\bm \alpha_{:,t} + \eta \bm \beta)^\top \tilde{\bm \theta} - \xi r_\lambda^2. 
\end{align*}
Since the objective function is a quadratic concave function
 w.r.t. $\eta$, we obtain the following by considering the condition (\ref{eq:KKT-eta}):
\begin{align*}
\eta = - \frac{\bm \alpha_{:,t}^\top \bm \beta}{\| \bm \beta \|^2_2}. 
\end{align*}
By substituting this into \eq{opt_theta},
\begin{align}
\label{opt_theta2}
\bm \theta = 
\tilde{\bm \theta} - \frac{1}{2\xi} \left(
\bm \alpha_{:, t} - \frac{\bm \alpha_{:,t}^\top \bm \beta}{\| \bm \beta \|^2_2} \bm \beta
\right). 
\end{align}
Since 
$\xi > 0$
and
\eq{complement}
indicates
$\| \bm \theta - \tilde{\bm \theta} \|^2_2 - r_\lambda^2 = 0$,
by substituting \eq{opt_theta2} into this equality,
\begin{align*}
\xi = \frac{1}{2 \| \bm \beta \|_2 r_\lambda} 
\sqrt{\| \bm \alpha_{:,t} \|^2_2 \| \bm \beta \|_2^2 - (\bm \alpha_{:,t}^\top \bm \beta)^2}. 
\end{align*}
Then,
from \eq{opt_theta2},
the solution of 
\eq{subprob2}
is given as
\begin{align*}
\bm \theta = 
\tilde{\bm \theta} -
\frac{\| \bm \beta \|_2 r_\lambda}
{\sqrt{\| \bm \alpha_{:,t} \|^2_2 \| \bm \beta \|_2^2 - (\bm \alpha_{:,t}^\top \bm \beta)^2}}
\left(
\bm \alpha_{:, t} - \frac{\bm \alpha_{:,t}^\top \bm \beta}{\| \bm \beta \|^2_2} \bm \beta
\right),
\end{align*}
and the minimum objective function value of 
\eq{subprob2}
is
\begin{align}
 \label{eq:lemma5-opt-objective-value}
\bm \alpha_{:,t}^\top \tilde{\bm \theta} - r_\lambda
\sqrt{\| \bm \alpha_{:,t} \|_2^2 - \frac{(\bm \alpha_{:,t}^\top \bm \beta)^2}{\| \bm \beta \|_2^2}}. 
\end{align}
Then, 
substituting 
\eq{eq:lemma5-opt-objective-value}
into 
\eq{eq:lemm5-decompose},
the optimal objective value of \eq{eq:ub}
is given as
\begin{align*}
\left| \bm \alpha_{:,t}^\top \tilde{\bm \theta} \right|
+ r_\lambda \sqrt{\| \bm \alpha_{:,t} \|_2^2 - \frac{(\bm \alpha_{:,t}^\top \bm \beta)^2}{\| \bm \beta \|_2^2}}. 
\end{align*}
\end{proof}

\paragraph{Proof of Lemma~\ref{lemm:4th}}
\begin{proof}
First,
using the bound introduced in \cite{kudo2004application}, 
\begin{align*}
\left| \sum_{i=1}^n \alpha_{it^\prime} \tilde{\theta}_i \right| &=
\left| \sum_{i: \beta_i \tilde{\theta}_i > 0} \alpha_{it^\prime} \tilde{\theta}_i + 
\sum_{i: \beta_i \tilde{\theta}_i < 0} \alpha_{it^\prime} \tilde{\theta}_i \right| \\
&\le \max \left\{
\sum_{i: \beta_i \tilde{\theta}_i > 0} \alpha_{it^\prime} \tilde{\theta}_i,~
-\sum_{i: \beta_i \tilde{\theta}_i < 0} \alpha_{it^\prime} \tilde{\theta}_i
\right\} \\
&\le \max \left\{
\sum_{i: \beta_i \tilde{\theta}_i > 0} \alpha_{it} \tilde{\theta}_i,~
-\sum_{i: \beta_i \tilde{\theta}_i < 0} \alpha_{it} \tilde{\theta}_i
\right\} \\
&=: u_t, 
\end{align*}
Next,
it is clear that 
\begin{align*}
\sum_{i=1}^n \alpha_{it^\prime}^2
 - \frac{(\sum_{i=1}^n \alpha_{it^\prime} \beta_i)^2}{\| \bm \beta\|_2^2}
\le \sum_{i=1}^n \alpha^2_{it^\prime}
\le \sum_{i=1}^n \alpha^2_{it} := v_t. 
\end{align*}
By combining them,
\begin{align*}
\left| \sum_{i=1}^n \alpha_{it^\prime} \tilde{\theta}_i \right| +
r_\lambda 
\sqrt{
\sum_{i=1}^n \alpha_{it^\prime}^2
 - \frac{(\sum_{i=1}^n \alpha_{it^\prime} \beta_i)^2}{\| \bm \beta\|_2^2}}
\le u_t + r_\lambda \sqrt{v_t}.
\end{align*}
\end{proof}